\documentclass[review]{elsarticle}

\usepackage{amsmath,amssymb,amsfonts,bm}
\usepackage{caption}
\usepackage{multirow}
\usepackage{subfigure}
\usepackage{booktabs}
\usepackage{algorithmic,algorithm}

\usepackage[colorlinks, linkcolor=blue, anchorcolor=blue, citecolor=blue]{hyperref}
\usepackage{geometry}
\journal{Journal}

\begin{document}
\captionsetup[figure]{labelfont={bf},labelformat={default},labelsep=period,name={Fig.}}

\begin{frontmatter}

\title{RecFNO: a resolution-invariant flow and heat field reconstruction method from sparse observations via Fourier neural operator}

\author[mymainaddress]{Xiaoyu Zhao}

\author[mymainaddress]{Xiaoqian Chen}

\author[mymainaddress]{Zhiqiang Gong \corref{cor1}}
\cortext[cor1]{Corresponding author: gongzhiqiang13@nudt.edu.cn}

\author[mymainaddress]{Weien Zhou}

\author[mymainaddress]{Wen Yao}

\author[mymainaddress]{Yunyang Zhang}

\address[mymainaddress]{Defense Innovation Institute, Chinese Academy of Military Science, No. 53, Fengtai East Street, Beijing 100071, China}

\begin{abstract}
Perception of the full state is an essential technology to support the monitoring, analysis, and design of physical systems, one of whose challenges is to recover global field from sparse observations.
Well-known for brilliant approximation ability, deep neural networks have been attractive to data-driven flow and heat field reconstruction studies.
However, limited by network structure, existing researches mostly learn the reconstruction mapping in finite-dimensional space and has poor transferability to variable resolution of outputs.  
In this paper, we extend the new paradigm of neural operator and propose an end-to-end physical field reconstruction method with both excellent performance and mesh transferability named RecFNO.
The proposed method aims to learn the mapping from sparse observations to flow and heat field in infinite-dimensional space, contributing to a more powerful nonlinear fitting capacity and resolution-invariant characteristic.
Firstly, according to different usage scenarios, we develop three types of embeddings to model the sparse observation inputs: MLP, mask, and Voronoi embedding.
The MLP embedding is propitious to more sparse input, while the others benefit from spatial information preservation and perform better with the increase of observation data.
Then, we adopt stacked Fourier layers to reconstruct physical field in Fourier space that regularizes the overall recovered field by Fourier modes superposition.  
Benefiting from the operator in infinite-dimensional space, the proposed method obtains remarkable accuracy and better resolution transferability among meshes. 
The experiments conducted on fluid mechanics and thermology problems show that the proposed method outperforms existing POD-based and CNN-based methods in most cases and has the capacity to achieve zero-shot super-resolution.
\end{abstract}

\begin{keyword}
Flow field reconstruction \sep Neural operator \sep Infinite-dimensional space \sep Sparse observation.
\end{keyword}

\end{frontmatter}

\section{Introduction}
\label{intro}
Understanding the system states is crucial for monitoring, controlling, analyzing, and designing the physical system, especially for applying digital twinning in natural systems.
One critical challenge of this technology is to reconstruct the global physical field (e.g. flow field and heat field) from sparse observations measured by a limited number of sensors \cite{callaham2019robust,carlberg2019recovering}. 
In particular, the reconstruction problem is usually ill-posed, and difficult to obtain appropriate results by directly solving the inverse problem. 
In general, data-driven methods are alternative approaches to address such issues by learning to recover unknown information from offline data, of which deep-learning based reconstruction method \cite{kutz2017deep} has made an impressive performance in recent years, benefiting from the well-known powerful approximate capacity of deep neural networks (DNNs).

Actually, the DNNs applied in field reconstruction mainly include model reduction-based \cite{wu2020data} and end-to-end methods.  
For the model reduction-based methods, the reconstruction task is explicitly decomposed into building reduced order model (ROM) and estimating low-dimensional coefficients.  
Earlier studies adopts traditional ROMs includes POD \cite{giannopoulos2020data,deng2019time} and DMD \cite{proctor2016dynamic} to generate reduced basis, the linear combination \cite{rowley2017model} of which is used to estimate the global field.
Then, the combination coefficient is usually obtained by solving an optimization problem \cite{everson1995karhunen,podvin2006reconstruction,mathelin2018observable} or estimated by regression model.
With the development of deep learning technology, researchers leveraged DNNs to build parameterized ROMs \cite{xu2020multi} for better dimensionality reduction, such as autoencoder \cite{wang2016auto,kumar2022state} and generative model \cite{creswell2018generative,morton2021parameter}, and establish deep regression model for high precision of coefficient estimation \cite{nair2020leveraging}.
Due to the powerful representational ability, these DNN methods can provide more competitive performance than traditional ROMs \cite{gruber2022comparison,dubois2022machine}. 
However, two stages of ROMs building and coefficients estimation will both introduce errors to influence reconstruction accuracy, and it has poor interpretability in conducting coefficients estimation.

In contrast with ROM-based method, this paper focuses on the end-to-end flow and heat field reconstruction method by deep neural networks. 
Existing studies to make direct reconstruction from sparse observations can be mainly divided into two classes: fully-connected neural network (FCNN) and deep convolutional neural network (CNN) architecture. 
Even though some studies \cite{erichson2020shallow} show FCNN presents better performance than POD-based methods in flow field reconstruction,  the low-efficiency problem exists in FCNN to make structured predictions by stacked deep neural networks.
In contrast, CNNs adopt parameters sharing and local connection to solve the low computational efficiency, and has been the state-of-the-art methods in various tasks \cite{long2015fully,gong2019cnn}. 
Therefore, deep CNNs have attracted more and more attention in physical field reconstruction studies. 
To leverage the CNN architecture, \cite{liu2021supervised} firstly used fully-connected layers to transform the low-dimensional observations into high-dimensional representation, which is then reshaped to 3D feature maps and taken as the inputs of residual CNN blocks to reconstruct physical field. 
\cite{gong2021physics} and \cite{peng4065493deep} adopted images to represent the position and measurements of sensors and developed image-to-image prediction architecture using CNN for temperature field reconstruction.
Though these methods have made important progress in field reconstruction of various physical systems, there are some issues limited by the network architecture. 
The physical field embodies the state of complex nonlinear system and has strong integrality governed by the implicit physical lows, but previous end-to-end methods have no global regularization like the referenced modes in POD-based method. It is promising to introduce function constraints in the overall field to improve generalization and accuracy of reconstruction methods.
Additionally, these methods are usually mesh-dependent that the trained model is limited by discretization in training data and shows poor transferability between different grid resolutions.

Recently, neural operator \cite{lu2019deeponet,nelsen2021random,li2020multipole} as a novel approximation paradigm of neural networks has showed their potentials to learn mesh-independent operator between infinite-dimensional spaces.
Unlike FCNN and CNN, neural operators introduce function space properties to constrain network training, demonstrating great nonlinear fitting and invariant discretization.
For example, \cite{li2020multipole} and \cite{anandkumar2020neural} composed nonlinear activation function and integral operators with graph neural network to learn discretization-invariant solution operators to partial differential equations (PDEs). 
Leveraging Fourier transform, \cite{li2020fourier} proposed Fourier neural operator to parameterize the integral operator in Fourier space. Numerical experiments show that Fourier neural operator outperforms most deep learning methods and achieves zero-shot super-resolution.
Currently, neural operator approaches are mainly applied to approximate PDE solutions with different functional parametric dependencies. 

Considering the merits of neural operator and deep learning, we extend the paradigm of neural operator to flow and heat field reconstruction and propose a novel end-to-end field reconstruction method named RecFNO, which aims to learn the mapping from sparse observations to physical field in infinite-dimensional space. By parameterizing the integral operator in Fourier space, the Fourier modes are superposed to produce the features, which regularize the predictions in spatial to improve generalization.
Besides, using the deep architecture with nonlinear activation function and parameterized integral operator can guarantee approximation ability of RecFNO.
Since the inputs of the problem at hand are sparse with less information compared with super-resolution task \cite{fukami2021machine,deng2019super}, we study the representations of inputs for RecFNO combining the works \cite{peng4065493deep,fukami2021global} for CNN architecture. 
Concretely, three types of embeddings are developed to represent the sparse observations for the proposed method, including MLP, mask, and Voronoi.
Mask and Voronoi embeddings preserve the spatial information of observation data, while MLP embedding is more flexible without prior information of sensor locations.
The size of three embeddings is adjustable to support the reconstruction results of different resolutions.
Finally, we conduct experiments in fluid mechanics and thermology cases, and the proposed representations show competitive results in different numbers of observations.     

In conclusion, it is critical to accurately reconstruct physical field from sparse observation as a challenge inverse problem. In contrast with existing methods using CNN and FCNN architectures, we propose a novel method to leverage global characters of Fourier modes to reconstruct the flow and heat field in infinite-dimensional space. Our method is also superior to POD-based methods, which are linear technologies and depends on the orthogonal basis under fixed resolution. The contributions of our work are as follows
\begin{enumerate}
	\item We propose a novel end-to-end method named RecFNO to reconstruct flow and heat field in Fourier space. Based on neural operator theory, the proposed method leverages function regularization to improve the reconstruction performance and has better resolution transferability between different meshes.
	\item We design three types of representations for RecFNO, which can adapt to different scenarios of learning from sparse observations. The experiments show that our method with different representations outputs competitive results, and appropriate representation should be selected for different situations.
	\item The numerical experiments demonstrate that the proposed methods outperform existing deep learning reconstruction approaches in most cases. In addition, our approach can achieve zero-shot super-resolution to provide higher resolution predictions than training data.
\end{enumerate} 

The remainder of this paper is organized as follows. The problem description of physical field reconstruction is presented in section 2. In section 3, the framework of the proposed methods is firstly described, then we introduce the different representations of sparse observations and the structure of Fourier layers. The comparison experiments in four numerical examples are performed and discussed in section 4. Finally, we conclude the proposed methods and experiment results in section 5.

\section{Problem description}
\label{sec:2}

Without loss of generality, the field of a physical system is described as follows:
\begin{equation}
	s=f(t;\lambda),
\end{equation}
where $s$ is the discretization of system state which is govern by the nonlinear function $f(\cdot)$ and depends on the parameters $\lambda \in \mathbb{R}^{n_\lambda}$ and time $t \in \mathbb{R}$. 
For the two-dimensional field $s \in \mathbb{R}^{n_y \times n_x}$ under $\lambda_i$ and $t_i$, sparse observations $o \in \mathbb{R}^n$ are obtained by measured at $n$ locations over the global field. 
The objective of our paper is to reconstruct the flow or heat field $s$ by leveraging the sparse observations $o$. 
It is a challenging inverse problem to restore unobserved information, and data-driven methods based on deep neural networks are developed in our paper to learn the mapping $g(\cdot)$ from the observations to the global field, which is expressed as
\begin{equation}
	s=g(o;\theta).
\end{equation}

In contrast with most existing method that attempts to learn the mapping $g(\cdot)$ from the finite-dimensional Euclidean spaces, we take advantage of the neural operator to reconstruct the field in function space.
Denoting the input function space $\mathcal{A}$ and output function space $\mathcal{U}$ defined in domain $D$, neural operator is developed to learn an infinite-dimensional-space mapping ${\mathcal{G}^\dag }:\mathcal{A} \to \mathcal{U}$ parameterized by the neural networks. 
Similar to the empirical-risk minimization problem in the finite-dimensional setting, the objective of training the neural operator $\mathcal{G}_{\theta}$ can be represented as 
\begin{equation}
	\mathop {\min }\limits_\theta  {\mathbb{E}_{a \sim p}}\left[ {C({\mathcal{G}_\theta }(a),{\mathcal{G}^\dag }(a))} \right],
\end{equation}
where $a$ is the function sampled from probability measure $p$ in $\mathcal{A}$, and $C:\mathcal{U} \times \mathcal{U} \to \mathbb{R}$ is the cost function between the function $u={\mathcal{G}^\dag }(a)$ and the predicted function $u'=\mathcal{G}_{\theta}(a)$.   
In general, discrete treatment is performed to represent the function $a(x)_i|_{D_i}\in \mathbb{R}^{n \times d_a}$ and $u(x)_i|_{D_i}\in \mathbb{R}^{n \times d_u}$ in $n$ positions $D_i = \left\{ {{x_1},...,{x_n}} \right\}$. 

Considering the above concepts, we aim to leverage the neural operator to learn the mapping from observations to the global field in infinite-dimensional spaces. For the reconstruction task at hand, discretization representation $a(x)_i|_{D_i}\in \mathbb{R}^{n \times d_a}$  of partial observations $o$ is generated as the initial function and we attempt to recover the full function $u(x)_i|_{D_i}\in \mathbb{R}^{n \times d_u}$ described the physical field $s$.

\section{Methods}
\label{sec:3}

Classical deep learning-based techniques for physical field reconstruction mostly learn the mapping from partial observations to global field in Euclidean spaces, which causes the method usually to be subject to the fixed discretization of training data. Inspired by the Fourier neural operator, we propose a competitive framework called RecFNO to reconstruct physical field in Fourier space. 
We aim to leverage the Fourier modes to regularize the reconstruction and introduce characteristics such as smoothness and resolution-independent in infinite dimensional to improve precision and transferability. 

\subsection{Framework}

\begin{figure*}[tp]
	\centering
	\includegraphics[width=1.0\linewidth]{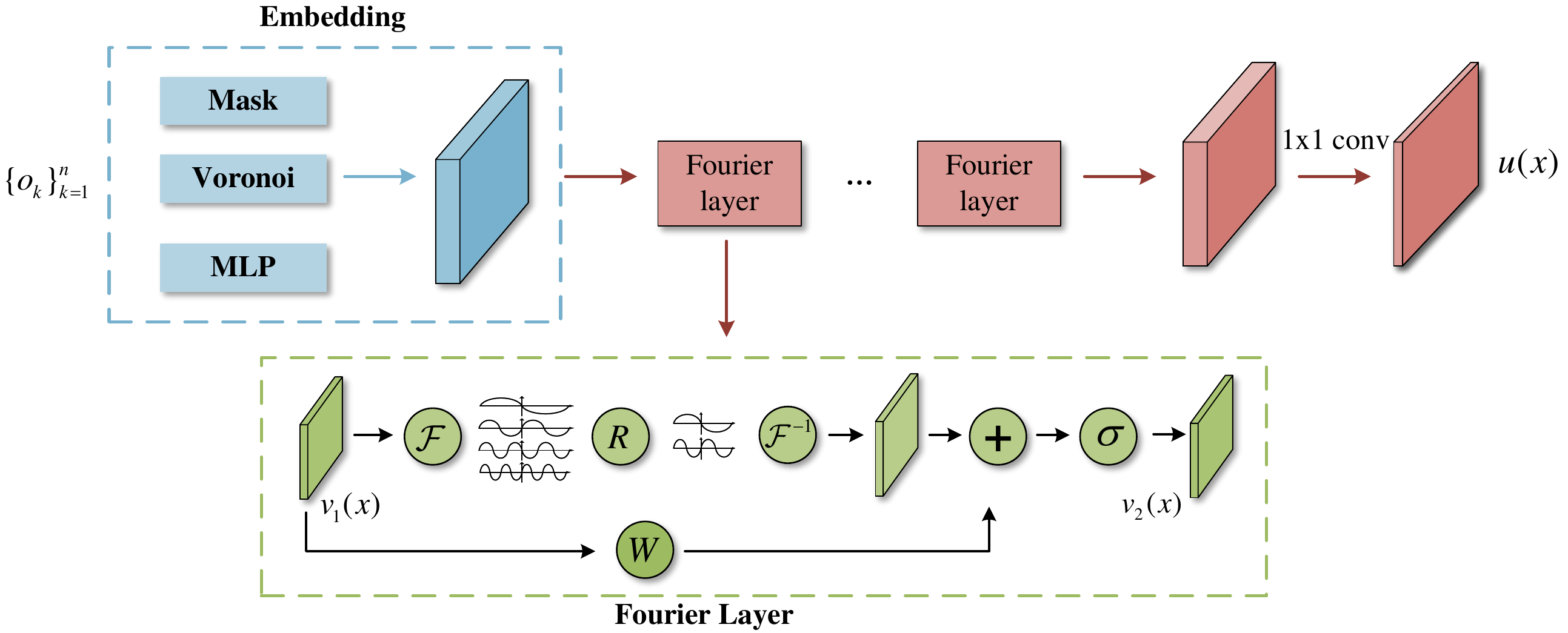}
	\caption{The pipeline of the proposed RecFNO. The RecFNO architecture is composed of an embedding module and multiple Fourier layers.}
	\label{fig:pipeline}
\end{figure*}

As shown in Fig.~\ref{fig:pipeline}, the proposed RecFNO framework learn the mapping from observations $\{ {o_k}\} _{k = 1}^n$ to the physical field $u(x)$. 
The RecFNO involves an embedding module and multiple Fourier layers. 
The embedding module provides a structural representation of sparse observations prepared for Fourier transform as the initial state in Fourier space. In this module, we investigate three categories of embedding, including mask, Voronoi, and MLP.
The positional information of sensors is required for mask and Voronoi embedding to preserve the spatial information. In contrast, MLP embedding is more flexible without prior location and has stronger representation ability with more sparse inputs.   
Additionally, embedding size determines the resolution of the prediction, and RecFNO can flexibly transfer to different meshes by adjusting the embedding size rather than limiting it by the fixed grid division of training data. 
Fast Fourier Transform (FFT) is conducted in each Fourier layer, then the non-linear mapping parameterized by the neural network is trained in the Fourier space. 
The stacked Fourier layers in the architecture improve the approximation ability to reconstruct physical fields more precisely.

\subsection{Embedding for sparse observations}

Compared with the super-resolution reconstruction problem, The inputs of task at hand are more sparse in spatial as they are the measurements on limited locations of global field. 
The representations of measurements are important for reconstructed models to capture enough information.
Based on the existing CNN-based architecture, we propose three manners of embedding for RecFNO to represent the sparse observations, consisting of mask, Voronoi \cite{fukami2021global}, and MLP embedding. 
Specifically, unknown parts in physical field are filled with zeros to obtain the mask representations of observations, and Voronoi is based on the mask to pad Unknown values with interpolation.
The mask and Voronoi preserve spatial information, but the position of the sensor needs to be known.
MLP embedding is expected to employ shallow full-connected neural networks for more flexible and powerful representations.
Different embeddings are suitable for different physical problems and settings that we conduct various experiments to analyze the applicability in numerical experiment section.

Without loss of generality, considering two-dimensional problems, the sparse observations are defined as $\{ ({x_k},{y_k}),{o_k}\} _{k = 1}^n$, which contains $n$ sensors positions $\{ ({x_k},{y_k})\} _{k = 1}^n$ and measurements $\{ {o_k}\} _{k = 1}^n$. Observations are projected to structured data as an image to the Fourier layers for the embedding.

\begin{figure*}[tp]
	\centering
	\includegraphics[width=0.75\linewidth]{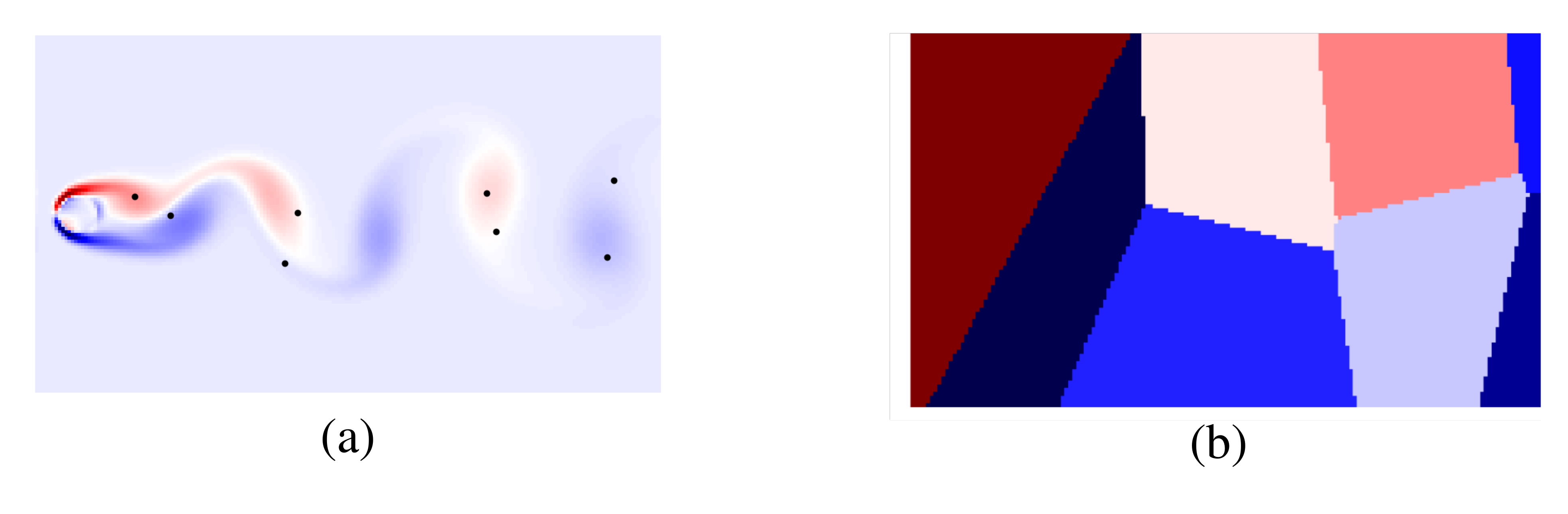}
	\caption{The representations of sparse observations.}
	\label{fig:representation}
\end{figure*}

\textbf{Mask embedding.} 
Discretizing the computational domain as matrix $D$ with size $(n_y, n_x)$, the coordinate in position  $(i, j)$ is $(D_{i,j}^x,D_{i,j}^y)$. To generate mask $M^{n_y \times n_x}$, measurements $\{ {o_i}\} _{i = 1}^n$ are placed on the positions  $\{ ({x_k},{y_k})\} _{i = 1}^n$ to reserve spatial relationship, and the rest positions are filled with zeros, which is described as
\begin{equation}
	{M_{i,j}} = \left\{ {\begin{array}{*{20}{l}}
		{{o_k}}&{{\rm{if~}}(D_{i,j}^x,D_{i,j}^y) = ({x_k},{y_k}){\rm{~for ~any~}}k}\\
		0&{{\rm{otherwise}}}
		\end{array}} \right..
\end{equation}

The mask $M$ is then concatenated with the coordinates $D_x$ and $D_y$ as the inputs of $1\times1$ convolution layer to generate the mask embedding ${\bf{e}}^{n_y \times n_x \times n_e}$, where $n_e$ is the length of feature vector in each location.   

\textbf{Voronoi embedding.} 
To pad observation values rather than zeros in all grid points, \cite{fukami2021global} proposed to project the observations $\{ {o_i}\} _{i = 1}^n$ on $n$ Voronoi tessellation, which is a partition of the spatial domain. Voronoi tessellation divide computational domain $D^{n_y \times n_x}$ into $n$ regions determined by the distance between grid points and observations. In each region $g_k$, specific observation $o_k$ is allocated as the center, which is the nearest observation of rest points in $g_k$. Therefore, the Voronoi representations $V$ is obtained by the nearest interpolation of $\{ ({x_k},{y_k}),{o_k}\} _{k = 1}^n$, which is described as
\begin{equation}
	{V_{i,j}} = {o_k},{\rm{if~}}d((D_{i,j}^x,D_{i,j}^y),({x_k},{y_k})) < d((D_{i,j}^x,D_{i,j}^y),({x_{k'}},{y_{k'}})),\forall k' \ne k,
\end{equation}
where $d(\cdot, \cdot)$ is distance function in Euclidean domain. In addition, 0-1 mask $M'$ is employed to identify the position of observations, defined as
\begin{equation}
{M_{i,j}} = \left\{ {\begin{array}{*{20}{l}}
	{{1}}&{{\rm{if~}}(D_{i,j}^x,D_{i,j}^y) = ({x_k},{y_k}){\rm{~for ~any~}}k}\\
	0&{{\rm{otherwise}}}
	\end{array}} \right..
\end{equation}

The example of mask and Voronoi representations is shown in Fig.~\ref{fig:representation}. Similar to mask embedding, the convolution layer takes the Voronoi image $V$, mask $M'$, coordinates $D_x$, and $D_y$ as inputs and produces the Voronoi embedding.
From the perspective of super-resolution, Voronoi representation is a low-resolution image of the physical field by interpolation, especially for the significant increase in the number of measurements. 
The Voronoi approach is robust to missing or moving measurements because the low-resolution image is adaptively generated according to position relationships. 

\textbf{MLP embedding.}
\begin{figure*}
	\centering
	\includegraphics[width=0.98\linewidth]{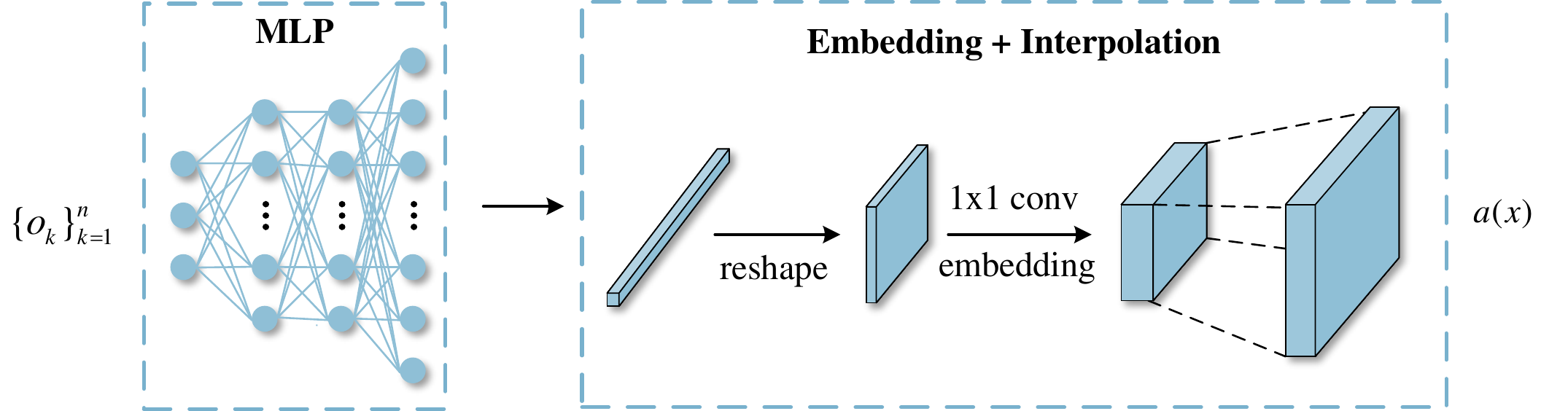}
	\caption{The illustrate of the MLP embedding.}
	\label{fig:mlp_emb}
\end{figure*}
The obtained representation by the above methods is sparse and unsmooth in the input map. To avoid roughly transforming the inputs to the function space, as shown in Fig.~\ref{fig:mlp_emb}, we study to leverage MLP and convolution layers to produce a relatively fine initial map for the sparse observations. 

Firstly, the MLP is employed to take the vanilla inputs $\{ {o_k}\} _{k = 1}^n$ to provide high-dimensional vectorized representations $g \in {\mathbb{R}^{n'}}$, where $n'=h' \times w'$. 
The MLP contains two layers with the GELU activation function. 
Furthermore, the vectore $g$ is reshaped into single-channel feature map $g' \in {\mathbb{R}^{h \times w \times 1}}$ and $1 \times1$ convolution layer is used to generate the embeddings ${\bf{e'}} \in {\mathbb{R}^{h \times w \times n_e}}$. 
In order to make the resolution of embedding the same as the output, the embeddings are resized to $(n_y, n_x)$, where $n_y$ are $n_x$ are the height and width of the output. 
In this paper, the embedding is firstly blown up to double size and processed by $3\times3$ convolution, and then expanded to  ${\bf{e}} \in {\mathbb{R}^{n_y \times n_x \times n_e}}$ by nearest neighbor interpolation.

\subsection{Fourier layer}
The stacked Fourier layers constitue an iterative architecture: ${v_0} \mapsto {v_1} \mapsto ... \mapsto {v_T}$ where $v_i$ is a function taking values in $\mathbb{R}^{d_v}$. 
In each Fourier layer, the functions update ${v_t} \mapsto {v_{t + 1}}$ is conducted by a kernel integral operator $\mathcal{K}$, a linear transformation $W$, and a non-linear activation function $\sigma$, which is expressed as 
\begin{equation}
	{v_{t + 1}}(x): = \sigma (W{v_t}(x) + (\mathcal{K}{(v_t)})(x)),
\end{equation}
\begin{equation}
	(\mathcal{K}({v_t}))(x) = \int {k(x,y){v_t}(y)d{v_t}(y)},
\end{equation}
where the kernel integral operator $\mathcal{K}$ and linear operator $W$ are bath parameterized and learnable. 
In each Fourier layer, Fourier transform is firstly performed, and parameterized kernel integral operator $\mathcal{K}$ is denoted as a convolution operator in Fourier space:
\begin{equation}
	(\mathcal{K}({v_t}))(x) = {\mathcal{F}^{ - 1}}(R \cdot \mathcal{F}({v_t}))(x),
\end{equation}
where $\mathcal{F}$ and $\mathcal{F}^{-1}$ define the Fourier transform and inverse Fourier transform, and $R$ is parameterization of a periodic function $k$ after Fourier transform. 

For the computational domain $D$, assuming the $v_t$ is discretized with a finite collection of observations, we have $v_t \in \mathbb{R}^{n \times d_v}$ and $\mathcal{F}({v_t}) \in \mathbb{C}^{n \times d_v}$. To improve the efficiency, the $k_{max}$ Fourier modes are supposed to characterize the $v_t$, which equals to low-pass filtering. The multiplication between the low-pass modes $\mathcal{F}({v_t}) \in \mathbb{C}^{k_{max} \times d_v}$ and the complex weight $R \in {\mathbb{C}^{{k_{\max }} \times {d_v} \times {d_v}}}$ is expressed as
\begin{equation}
	{(R \cdot \mathcal{F}({v_t}))_{k,i}} = \sum\limits_{j = 1}^{d_v} {{R_{k,i,j}}{{(\mathcal{F}({v_t}))}_{k,j}},\forall k = 1,...,{k_{\max }},i = 1,...,d_v}
\end{equation}

In this paper, the cases with two-dimensional domain and uniform discretization are investigated, and $\mathcal{F}$ and  $\mathcal{F}^{-1}$ can be implemented with the efficient FFT and Inverse Fast Fourier Transform (IFFT). 
For the inputs with the resolution $h \times w$ , the FFT $\mathcal{\hat F}$ and IFFT $\mathcal{\hat F^{-1}}$ are calculated as follows:
\begin{equation}
	(\mathcal{\hat F}{v_t})({k_1},{k_2}) = \sum\limits_{{x_1} = 0}^{h - 1} {\sum\limits_{{x_2} = 0}^{w - 1} {{v_t}({x_1},{x_2}){e^{ - 2i\pi (\frac{{{x_1}{k_1}}}{h} + \frac{{{x_2}{k_2}}}{w})}}} },
\end{equation}
\begin{equation}
	({\mathcal{\hat F}^{ - 1}}{v_t})({x_1},{x_2}) = \sum\limits_{{k_1} = 0}^{h - 1} {\sum\limits_{{k_2} = 0}^{w - 1} {{v_t}({k_1},{k_2}){e^{2i\pi (\frac{{{x_1}{k_1}}}{h} + \frac{{{x_2}{k_2}}}{w})}}} },
\end{equation}
where $(k_1,k_2) \in \mathbb{Z}_w \times \mathbb{Z}_h$, and $(x_1,x_2)\in D$. For $v_t \in \mathbb{R}^{d_v}$ and $v_{t+1} \in \mathbb{R}^{d_v}$, when the maximum number of modes are set to $({k_1}_{max}, {k_2}_{max})$, the $R$ is parameterized as the complex values $R \in \mathbb{C}^{{k_1}_{max}, {k_2}_{max},d_v,d_v}$.  
Additionally, denoting the discretization of $v_t \in \mathbb{R}^{d_v}$ with the resolution $h \times w$ as multi-channel feature map $\mathbb{R}^{h \times w \times d_v}$, the linear transformation $W \in \mathbb{R}^{d_v \times d_v}$ can be conducted by $1\times1$ convolution.

In the architecture of RecFNO, multiple Fourier layers are stacked to improve the representational capacity of the framework. 
Though the low-pass filtering performs in the kernel limits the number of Fourier modes, the multilayered structure with nonlinear activation function can considerably approximate operator mappings between function spaces. 

\subsection{Zero-shot super-resolution reconstruction}
The proposed reconstructed architecture is able to achieve zero-shot super-resolution, benefiting from the discretization-invariant of Fourier layer. 
Leveraging the Fourier transform, the parametric mapping is learned in the Fourier space, and the output of each Fourier layer is the combination of Fourier modes. 
Therefore, the Fourier layer can present different resolution results according to the discretization of the computation domain. 
In our architecture, the embedding size determines the discretization of the initial functions that feed into the Fourier layer to produce outputs with the same resolution. 
Therefore, the super-resolution results can be obtained by refining the grid of inputs. In this paper, we perform nearest neighbor interpolation to increase the resolution of embedding corresponding to the refined grid.
For example, assuming that the model is trained under the size of $(h,w)$, the double super-resolution is achieved when the embedding size is adjusted to $(2 \times h,2 \times w)$ without retraining the model.

\section{Numerical experiments}
\label{sec:4}

\subsection{Datasets and experiment settings}
\label{sec:4.1}

In this section, four datasets are presented to investigate the performance of the proposed approaches, including fluid mechanics benchmarks, 2D steady state heat conduction, and realistic geophysical problem. 
More details of these datasets are described as follows.

\begin{figure*}
	\centering
	\includegraphics[width=0.95\linewidth]{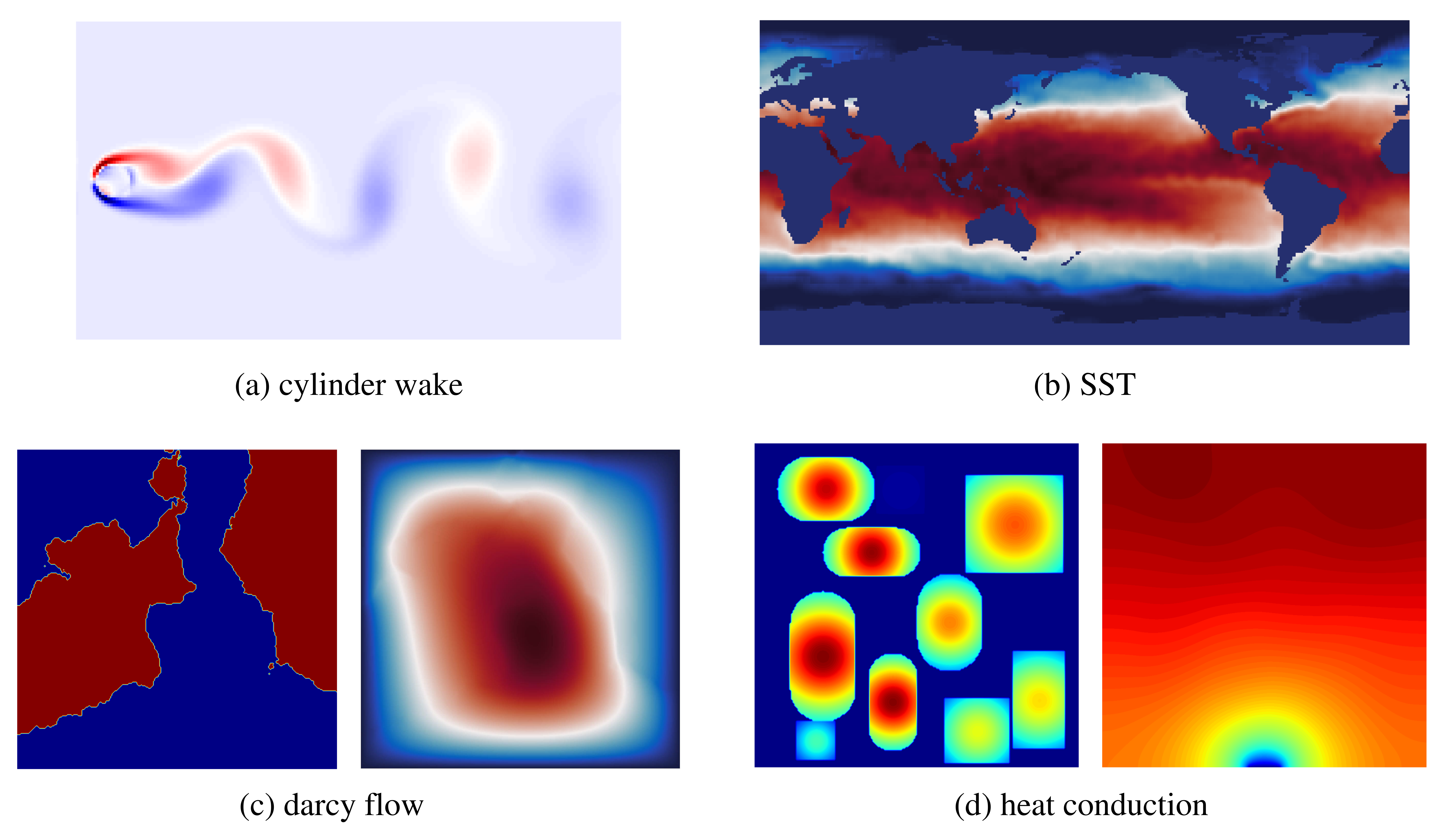}
	\caption{Illustrations of four datasets used in this paper.}
	\label{fig:dataset}
\end{figure*}

\noindent \textbf{2D Cylinder Wake.} 
For the first example, we consider reconstructing the vorticity in the two-dimensional unsteady cylinder wake, which is a well-known benchmark in the mechanic's community. 
Specifically, a cylinder body is located in the rectangular computational domain, whose flow field is characterized by periodic laminar flow vortex shedding. 
The simulation data can be obtained to solve the incompressible Navier-Stokes equations by numerical method. 
In particular, \cite{fukami2021global} provided a public cylinder wake dataset containing 5000 snapshots and spanning four vortex shedding periods. 
In preparing data, the Reynolds number is set to 100, and the size of each snapshot is (112, 192). In our experiments, the dataset is split into 3500, 750, and 750 in a timed sequence for training, validation, and testing.

\noindent \textbf{2D Steady-state Darcy Flow.} 
The 2D steady-state Darcy flow is considered as the second example. 
The dataset of Darcy flow in this paper is generated by solving the following elliptic PDE:
\begin{equation}
	\begin{array}{*{20}{l}}
	{{\rm{ - }}\nabla (a(x)\nabla u(x)) = f(x)}&{x \in \Omega },\\
	{u(x) = 0}&{x \in \partial \Omega },
	\end{array}
\end{equation}
where $\Omega$ is $[0,1]\times[0,1]$ square domain with the Dirichlet boundary $\partial \Omega$. 
The source item $f$ is identity function in the computation domain, while different diffusion coefficient functions $a$ is sampled to produce physical fields $u$ with the resolution of $(128, 128)$.
We assume the diffusion coefficient distribution is unknown to reconstruct global physical field from the measurements. 
Subsequently, the dataset containing 4000 training data, 1000 validation data, and 1000 test data is generated.

\noindent \textbf{Sea Surface Temperature Dataset.}
The sea surface temperature (SST) dataset is the realistic observation data collected and released by the National Oceanic and Atmospheric Administration (NOAA). 
The commonly used SST dataset comprises the weekly global sea surface temperature measurements from 1981 to 2018. 
It contains 1914 snapshots, and each snapshot is $180\times360$ gridded data covering global region. 
The initial 1500 snapshots are selected as the training dataset, and the following 200 and 214 snapshots are regarded as the validation and test dataset, respectively.

\noindent \textbf{2D Steady-state Heat Conduction.}
As the last example, we consider field reconstruction for the 2D non-linear, steady-state heat conduction problem \cite{gong2021physics,zhao2021surrogate}. 
There are some Gaussian heat sources placed on the $[0,0.1]\times[0,0.1]$ aquare domain $\Omega$. 
The generated heat is dissipated through the bottom thermostatic sink $\Omega _D$. The governing equation can be expressed as follow.
\begin{equation}
	\begin{array}{*{20}{l}}
	{{\rm{ - }}\nabla (\lambda \nabla u(x)) = f(x)}&{x \in \Omega },\\
	{u(x) = {u_D}}&{x \in \partial {\Omega _D}},\\
	{\lambda \frac{{\partial u}}{{\partial {\bf{n}}}} = 0}&{x \in \partial {\Omega _N}},
	\end{array}
\end{equation}
where ${\partial {\Omega _D}}$ and ${\partial {\Omega _N}}$ denote the Dirichlet and Neumann Boundary. 
$\lambda$ is the conductivity coefficient that varies with temperature $u$ as $\lambda=1+0.05\times(u-298)$. 
We sample the heat $u_D$ in sink and the intensity distribution $f(x)$ to generate various steady-state temperature field, in which 4000, 1000, and 1000 data are randomly chosen as training, validation, and test dataset.

\noindent \textbf{Evaluation Metrics.} 
The mean absolute error (MAE) $\varepsilon  = \left\| {{\bf{u}} - {{\bf{u}}_p}} \right\|/\left| {\bf{u}} \right|$ is chosen as the quantitative metric to measure the gap between the predicted field ${{{\bf{u}}_p}}$ and the referenced field ${\bf{u}}$, where ${\left| {\bf{u}} \right|}$denotes the total number of points in field $\bf{u}$. 
Additionally, we report the max absolute error (Max-AE) \cite{chen2021deep} ${\varepsilon _{\max }} = {\left\| {{\bf{u}} - {{\bf{u}}_p}} \right\|_\infty }$ to evaluate the prediction performance in the worst situation.

\noindent \textbf{Implementation Details.} 
In the experiments, the total epochs and batch size are set to 300 and 16 for cylinder wake and SST dataset, 200 and 8 for Darcy flow and heat conduction dataset. 
Adam optimizer with an initial learning rate of 0.001 is employed, and the learning rate is adjusted with a multiplicative factor of 0.98 (300 epochs) or 0.97 (200 epochs) after each epoch. 
We implement all models using Pytorch framework, and the models are trained and evaluated on a single Tesla V100. The implementation of all approaches and datasets are released at \url{https://github.com/zhaoxiaoyu1995/RecFNO}.

\subsection{Results}
\subsubsection{Results of comparison experiments}
To illustrate the superiority of the RecFNO architecture, we compare the proposed method with MLP-POD and CNN-based methods. 
MLP-POD is a classical and efficient reconstruction method that uses proper orthogonal decomposition (POD) to decompose each snapshot into POD modes and corresponding coefficients. 
Sequentially, the MLP is trained to learn the mapping from the observations to the coefficients of principal modes, combining with the principal modes to reconstruct the global physical field.
Benefiting from parameters sharing and partially connecting, CNN architecture has been demonstrated to reconstruct physical field with accuracy and efficiency \cite{peng4065493deep,fukami2021global,morimoto2022generalization}.
Therefore, the results obtained using CNN architecture are compared with the proposed method.
In comparison experiments, considering the small size of feature maps generated by MLP embedding, the Fourier layer is replaced with $3\times3$ convolution layers and $2\times$ interpolation to produce reconstructed results with original size.
As a fully convolutional neural network, UNet has been employed for various image-to-image tasks and proved to predict physical field with excellent performance.
Therefore, we compare the results of UNet and FNO architecture for mask and Voronoi embedding, which is the image-like data with the same size of predictions.

\begin{table}[htbp]
	\centering
	\caption{Comparison results of MLP-POD, CNN, and FNO architecture in four datasets. The second row presents two settings with different numbers of sensors on each dataset. For CNN and FNO architecture, the results of MLP, mask, and Voronoi embedding are presented. The MAE metric evaluates the reconstruction performance.}
	\begin{tabular}{cccccccccc}
		\hline\noalign{\smallskip}
		\multirow{2}[0]{*}{Embedding} & \multirow{2}[0]{*}{Model} & \multicolumn{2}{c}{Cylinder (e-4)} & \multicolumn{2}{c}{Darcy (e-4)} & \multicolumn{2}{c}{SST (e-1)} & \multicolumn{2}{c}{Heat (e-2)} \\
		\cline{3-10}
		&       & 2     & 4     & 16    & 25    & 64    & 128   & 25    & 36 \\
		\noalign{\smallskip} \hline \noalign{\smallskip}
		—     & MLP-POD & 5.314  & 3.065  & 2.749  & 2.739  & 3.364  & 3.288  & 7.992  & 8.289  \\
		\noalign{\smallskip} \hline \noalign{\smallskip}
		\multirow{2}[0]{*}{MLP} & CNN   & 1.750  & 1.591  & 0.928  & \textbf{0.682}  & 3.172  & 2.870  & 1.982  & 2.005  \\
		& FNO   & \textbf{0.567}  & \textbf{0.488}  & 1.041  & 0.736  & 3.419  & 3.116  & 0.438  & 0.414  \\
		\noalign{\smallskip} \hline \noalign{\smallskip}
		\multirow{2}[0]{*}{Mask} & UNet  & 2.167  & 2.129  & \textbf{0.855}  & 0.719  & 3.093  & 2.734  & 0.583  & 0.367  \\
		& FNO   & 3.364  & 1.457  & 0.977  & 0.772  & 3.076  & 2.777  & 2.988  & 1.126  \\
		\noalign{\smallskip} \hline \noalign{\smallskip}
		\multirow{2}[0]{*}{Voronoi} & UNet  & 2.241  & 2.445  & 1.154  & 0.836  & \textbf{3.037}  & 2.712  & 0.488  & 0.465  \\
		& FNO   & 0.780  & 0.540  & 0.980  & 0.744  & 3.197  & \textbf{2.706}  & \textbf{0.377}  & \textbf{0.294}  \\
		\noalign{\smallskip}\hline
	\end{tabular}%
	\label{tab:results}%
\end{table}%

Table~\ref{tab:results} presents the reconstruction precision with MLP-POD, CNN, and FNO architecture on four datasets. The performance of different methods is evaluated by MAE metric, and two settings with different numbers of sensors are considered for each dataset. 
From the results, as the end-to-end reconstruction architecture with deep neural networks, CNN and FNO-based methods outperform MLP-POD on all four datasets. 
In particular to the cylinder dataset with two observations, FNO with MLP embedding improves the precision by an order of magnitude than MLP-POD.
From results shown in Table~\ref{tab:results}, the type of embedding has a considerable impact on the reconstruction performance, but it is relevant to the problem, model architecture, and the number of sensors.
Specifically, the performance of MLP and Voronoi embedding is more stable than mask embedding. MLP embedding outperforms Voronoi for the cylinder problem with few observation values, while Voronoi is better for SST and Heat problems with more observation values. Mask embedding performs better in CNN than FNO architecture, as the reconstruction precision with FNO decreases remarkably on the cylinder and heat datasets.

\begin{figure*}[tp]
	\centering
	\subfigure[cylinder wake with 4 sensors]{
		\includegraphics[width=0.40\linewidth]{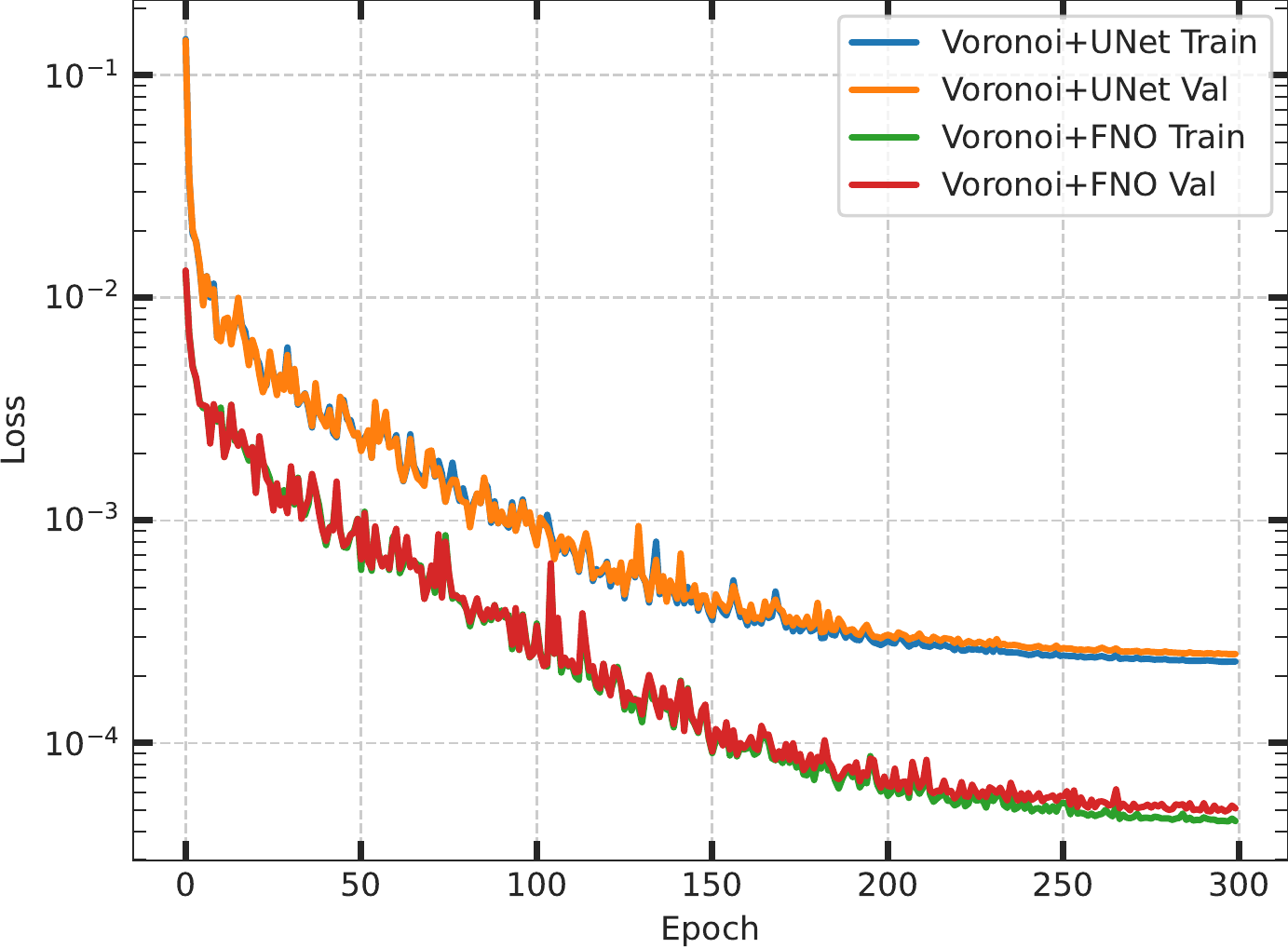}
	}
	\subfigure[darcy flow with 25 sensors]{
		\includegraphics[width=0.40\linewidth]{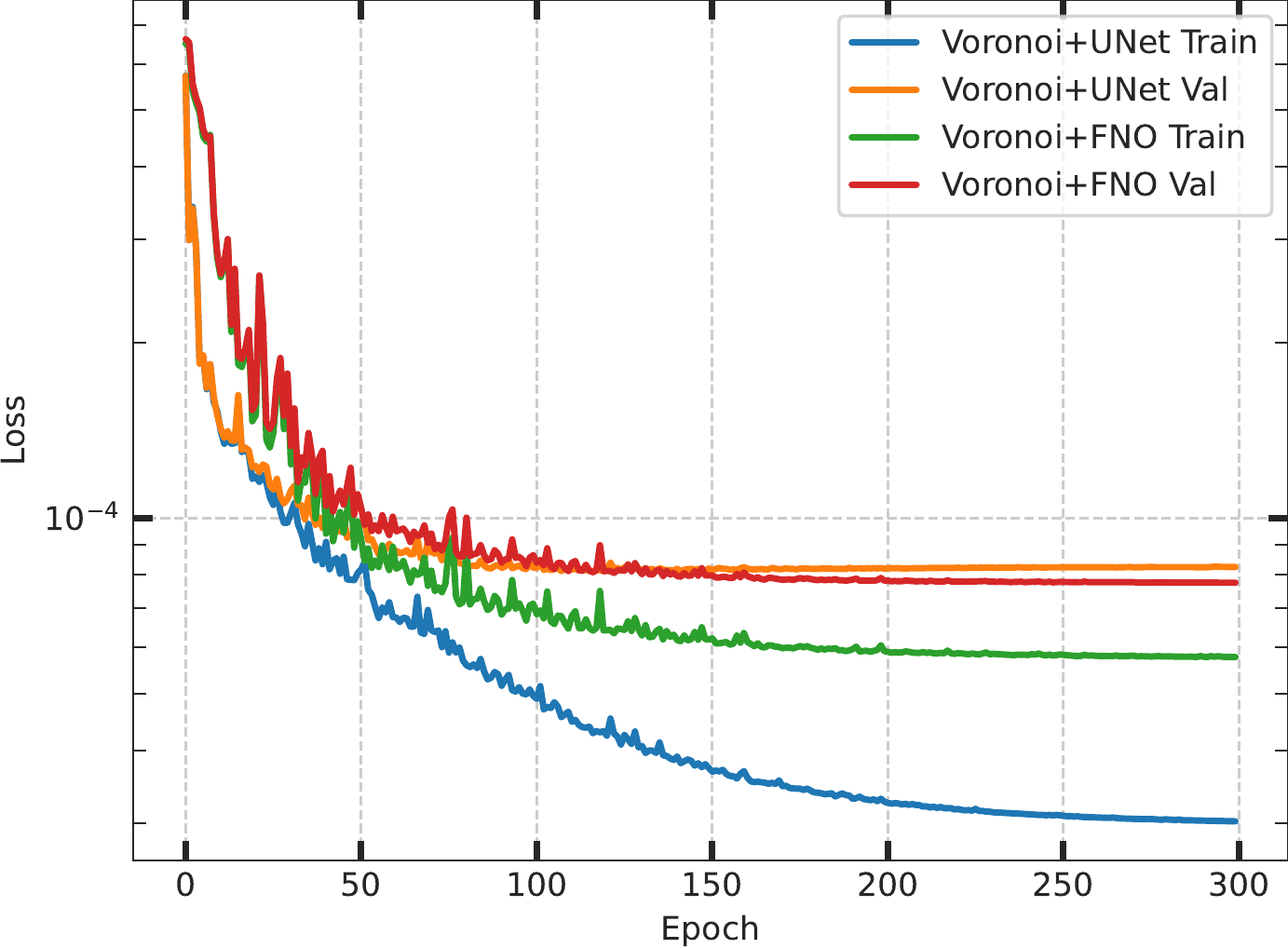}
	}
	\caption{Learning history of UNet and FNO with Voronoi embedding. The average L1 loss on training and validation sets are reported.}
	\label{fig:loss}
\end{figure*}

From the perspective of model architecture, FNO significantly improves the accuracy of field reconstruction on cylinder and heat datasets compared with CNN architecture and yields competitive results on Darcy and SST datasets.
In particular, the training and validation loss of FNO and UNet using the same Voronoi embedding are reported in Fig.~\ref{fig:loss}. For the cylinder wake with 4 sensors, FNO architecture shows a more powerful approximation ability, and the average L1 loss in the training and validation set is obviously lower than UNet. UNet obtains a smaller training loss on Darcy flow problem, but it suffers from severe overfitting that the margin of validation loss between FNO and UNet is narrow since the features are transformed in the Fourier space, which imposes regularization in the physical field reconstruction for FNO.     
Overall, the results demonstrate the applicability and efficiency of the proposed method, which employs the Fourier neural operator to reconstruct physical field in function space. 

\subsubsection{Results under different number of sensors}
\begin{figure*}[t]
	\centering
	\subfigure[cylinder wake]{
		\includegraphics[width=0.40\linewidth]{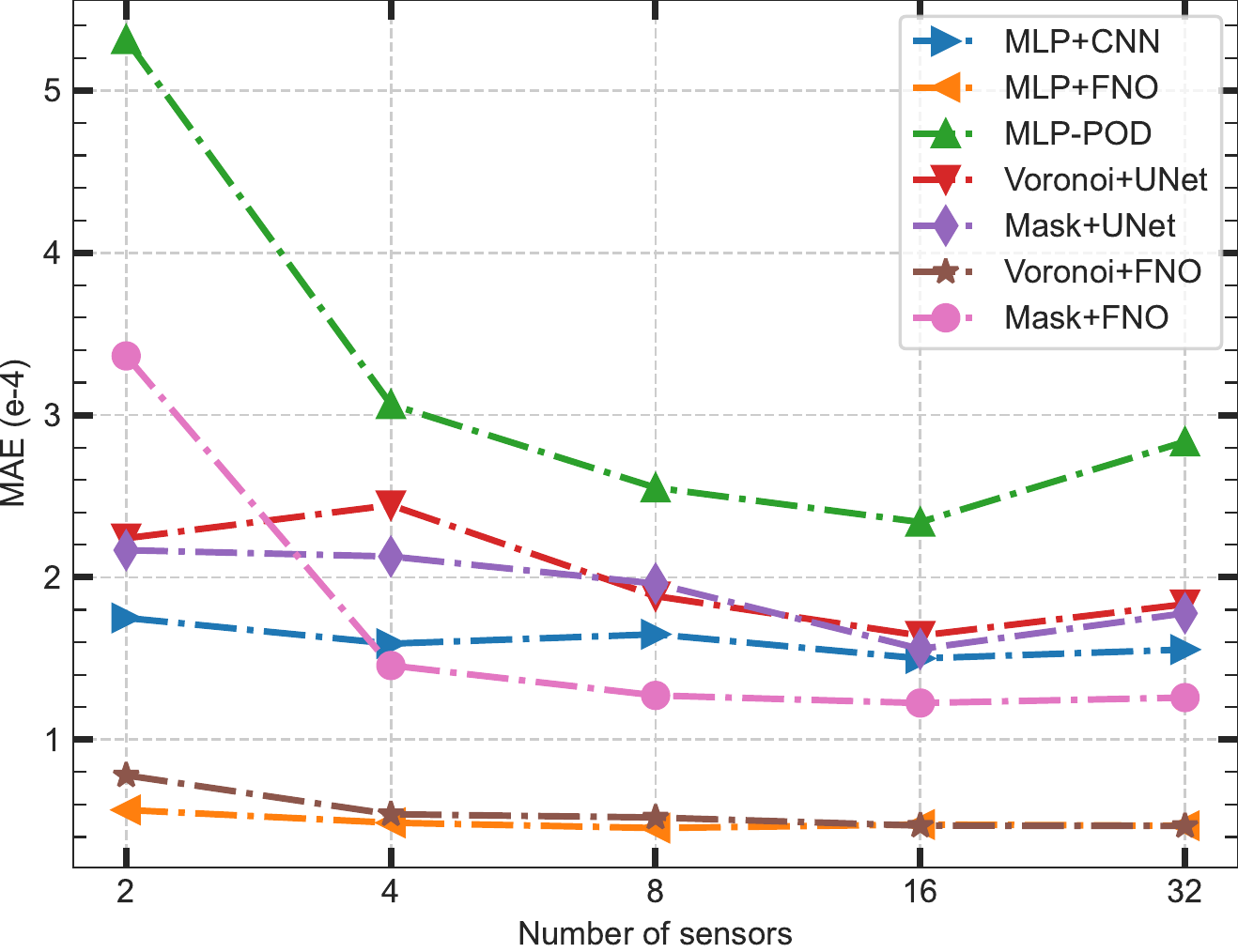}
	}
	\subfigure[Darcy flow]{
		\includegraphics[width=0.41\linewidth]{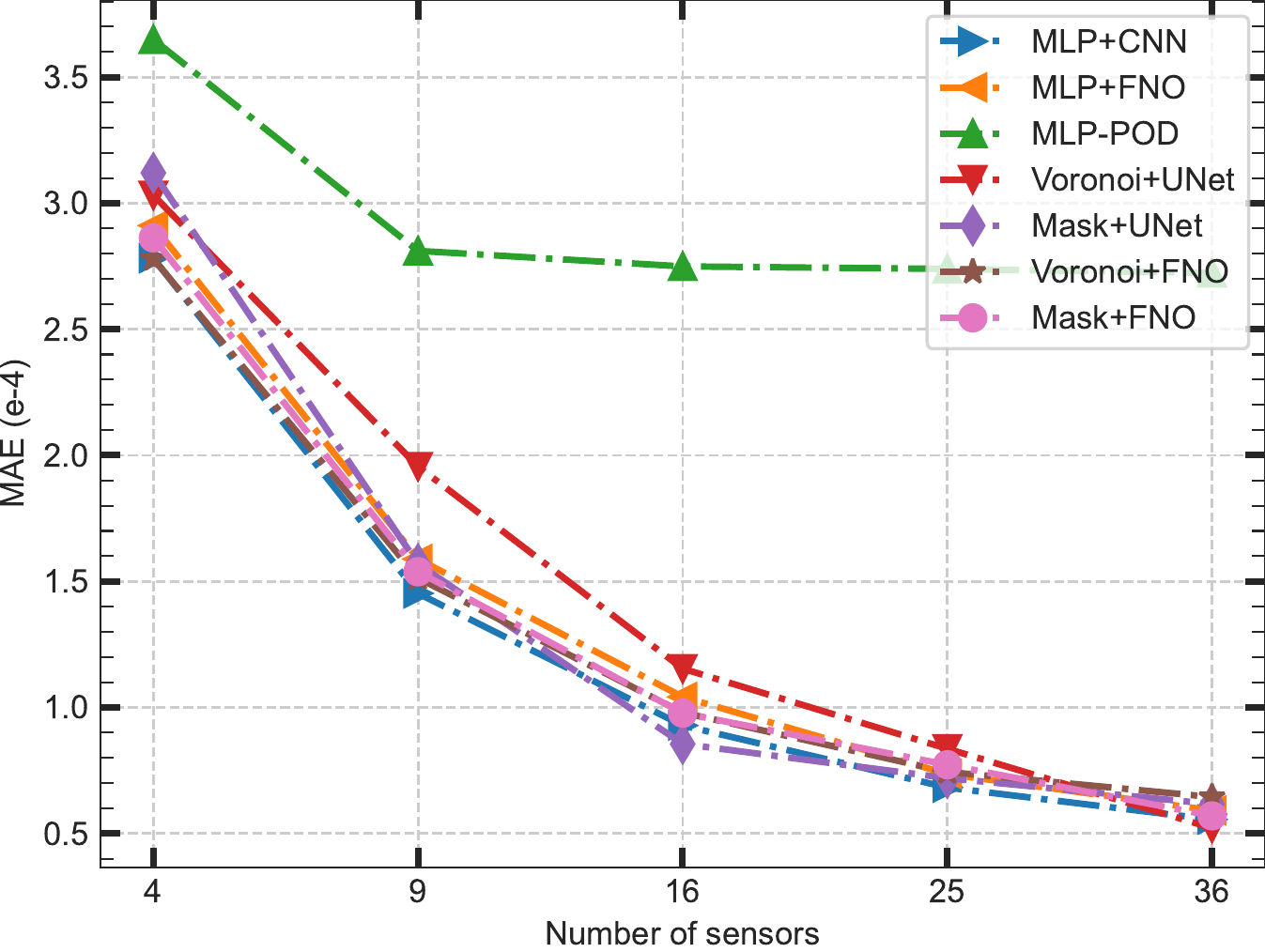}
	} \\
	\subfigure[sea surface temperature]{
		\includegraphics[width=0.40\linewidth]{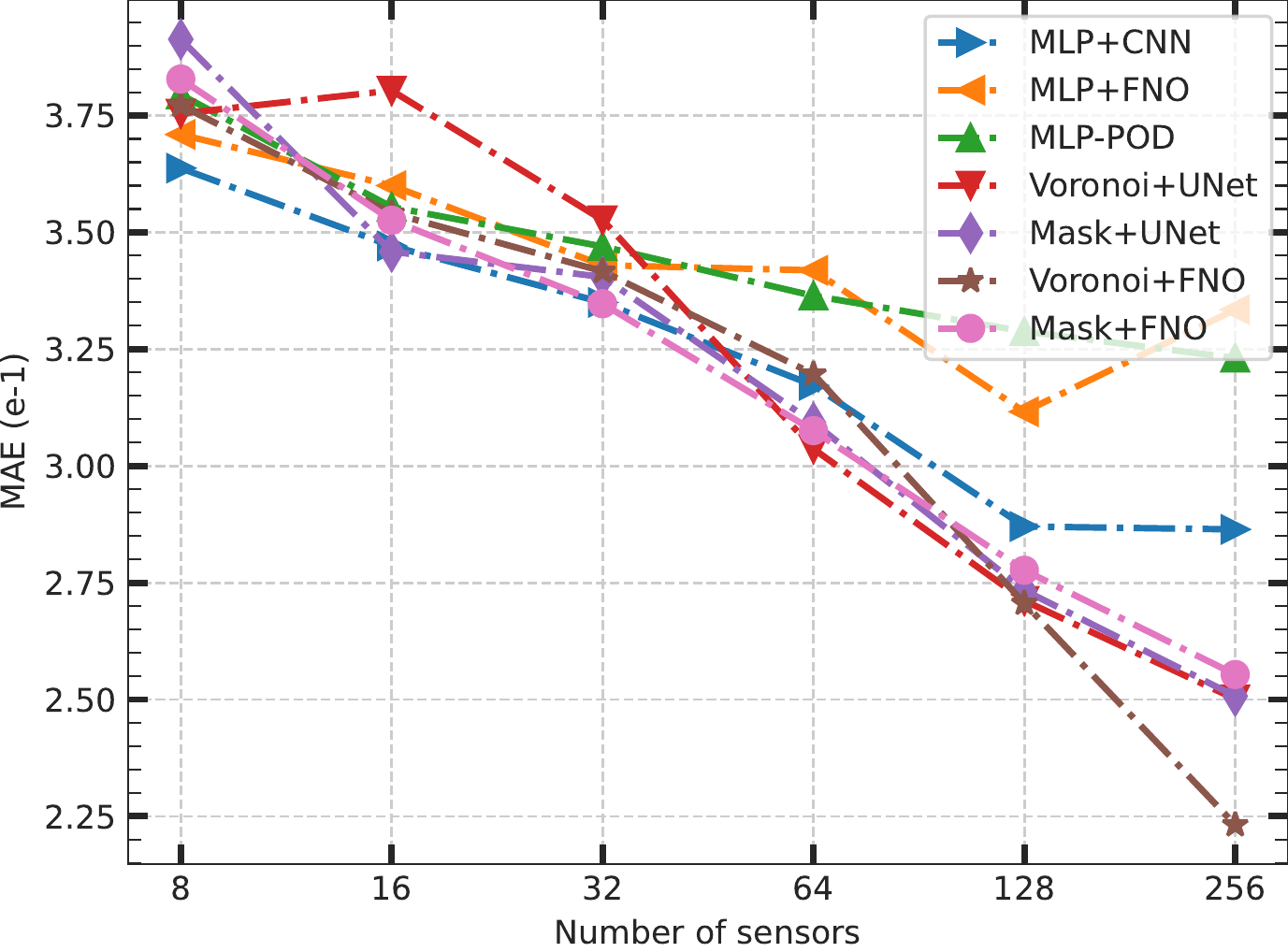}
	}
	\subfigure[heat conduction]{
		\includegraphics[width=0.41\linewidth]{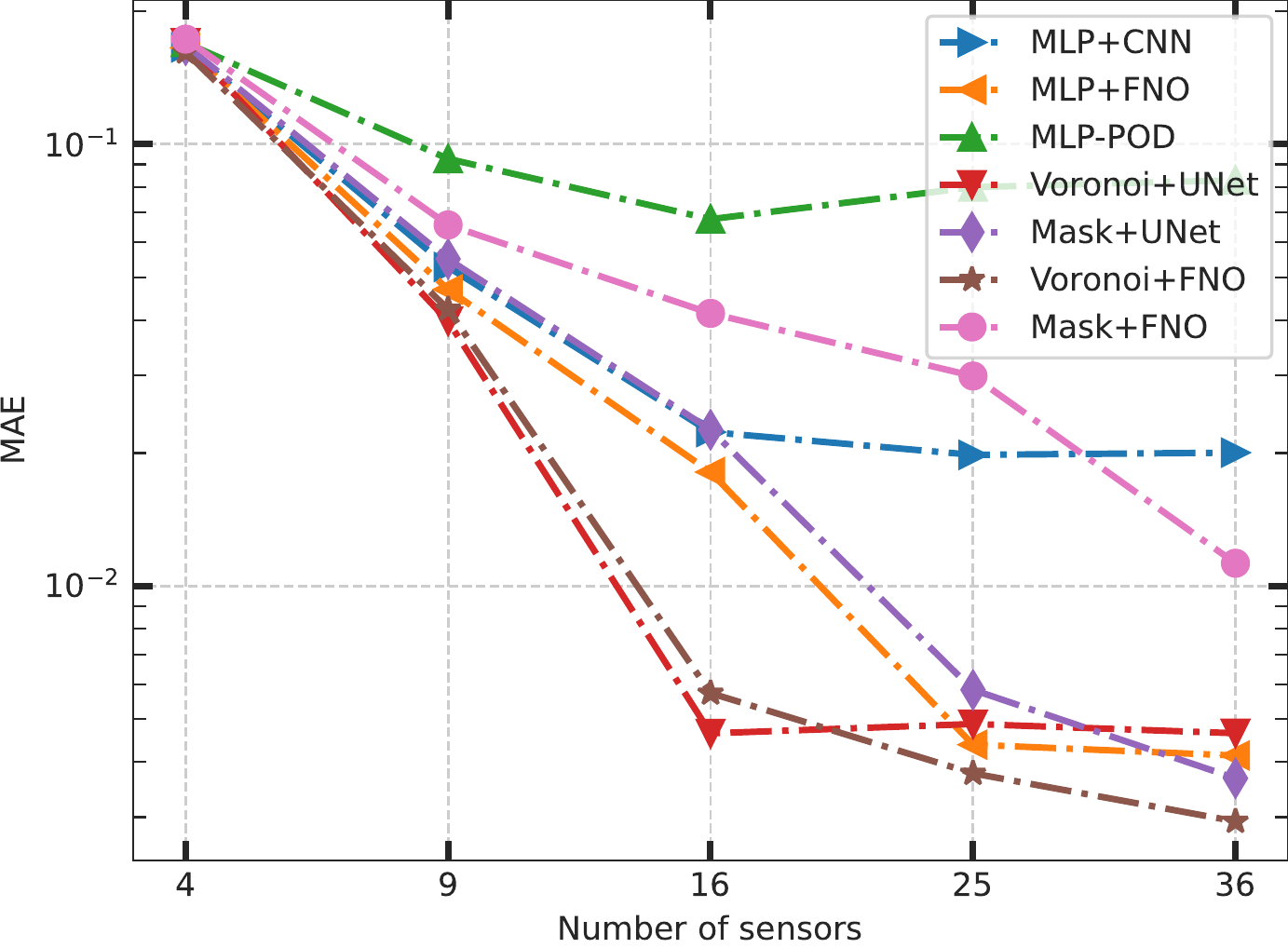}
	}
	\caption{Comparison of different approaches with an increasing number of sensors. All results are evaluated on MAE metric.}
	\label{fig:mae}
\end{figure*}

It is ideal to obtain enough observations to improve reconstruction accuracy, but specific application will limit the number of placed sensors. 
Therefore, we investigate the effect of varying numbers of observations.
Fig.~\ref{fig:mae} shows the MAE metric of different approaches with an increasing number of sensors on four datasets.
Adopting the MLP and Voronoi embedding, the proposed RecFNO reconstruction architecture yields the best results in most cases. When the number of sensors is very limited, the performance of MLP embedding is more stable. This is because the fully-connected neural network extracts information from observations to produce preliminary maps, which is more efficient than directly processing the sparse representations of mask and Voronoi. 
However, MLP embedding ignores the spatial information of observations. Therefore, Voronoi embedding shows better reconstruction performance with an increasing number of sensors, especially combining with the FNO architecture. 
The four cases show the different trends of model performance.   
The performance gains slightly to increase the number of sensors for cylinder wake problem, and four observations are able to reconstruct the vorticity precisely.
For the other datasets, especially the sea surface temperature, different models achieve significant performance improvement with the increase in sensor number.
The MLP-POD method follows different trend, which is not sensitive to the increase in sensor number.
As an end-to-end method using deep neural network, the results show that the proposed methods have a larger capacity to reconstruct physical field accurately.

\begin{figure*}[tp]
	\centering
	\subfigure[cylinder wake]{
		\includegraphics[width=0.40\linewidth]{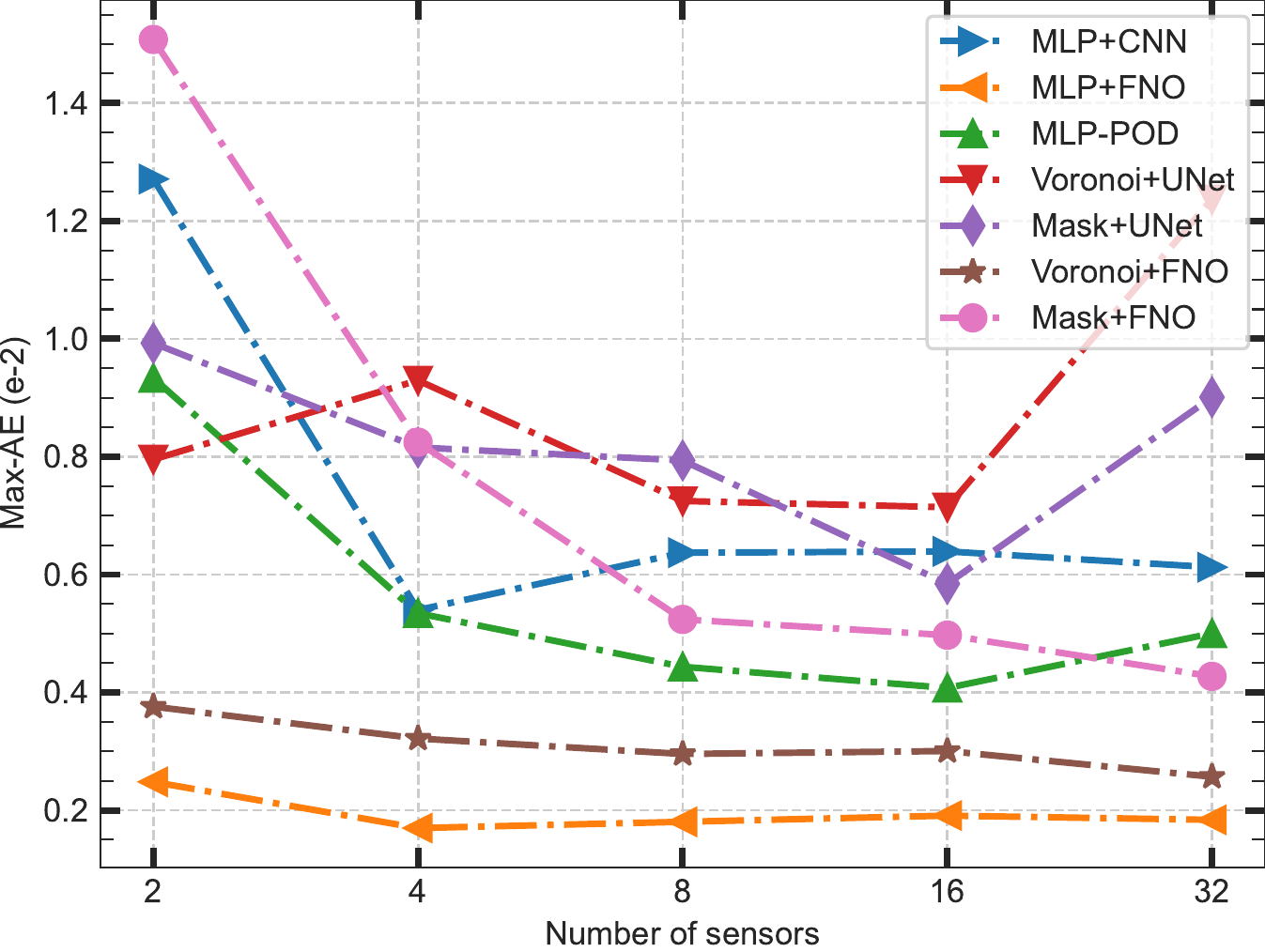}
	}
	\subfigure[darcy flow]{
		\includegraphics[width=0.405\linewidth]{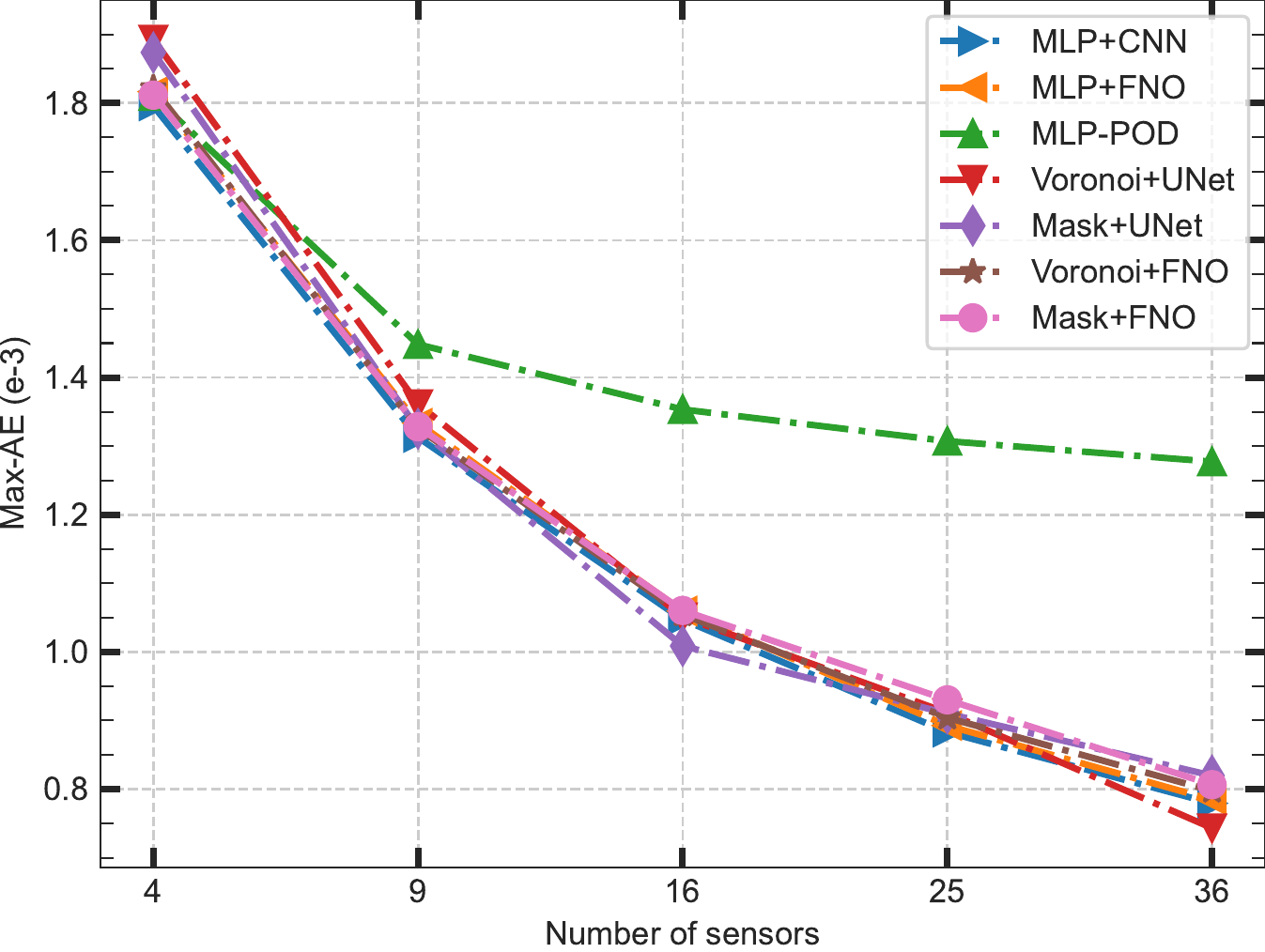}
	} \\
	\subfigure[sea surface temperature]{
		\includegraphics[width=0.40\linewidth]{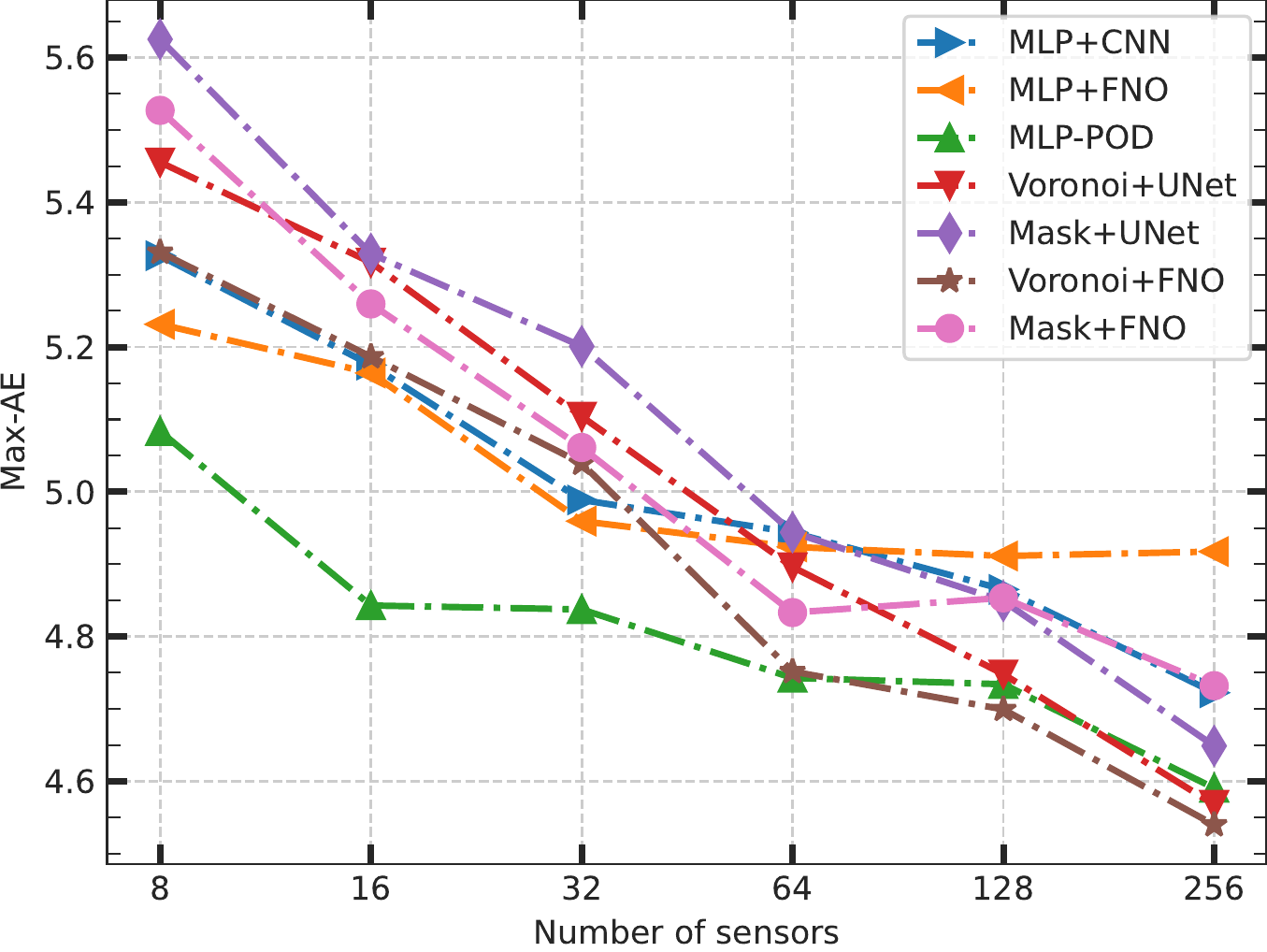}
	}
	\subfigure[heat conduction]{
		\includegraphics[width=0.405\linewidth]{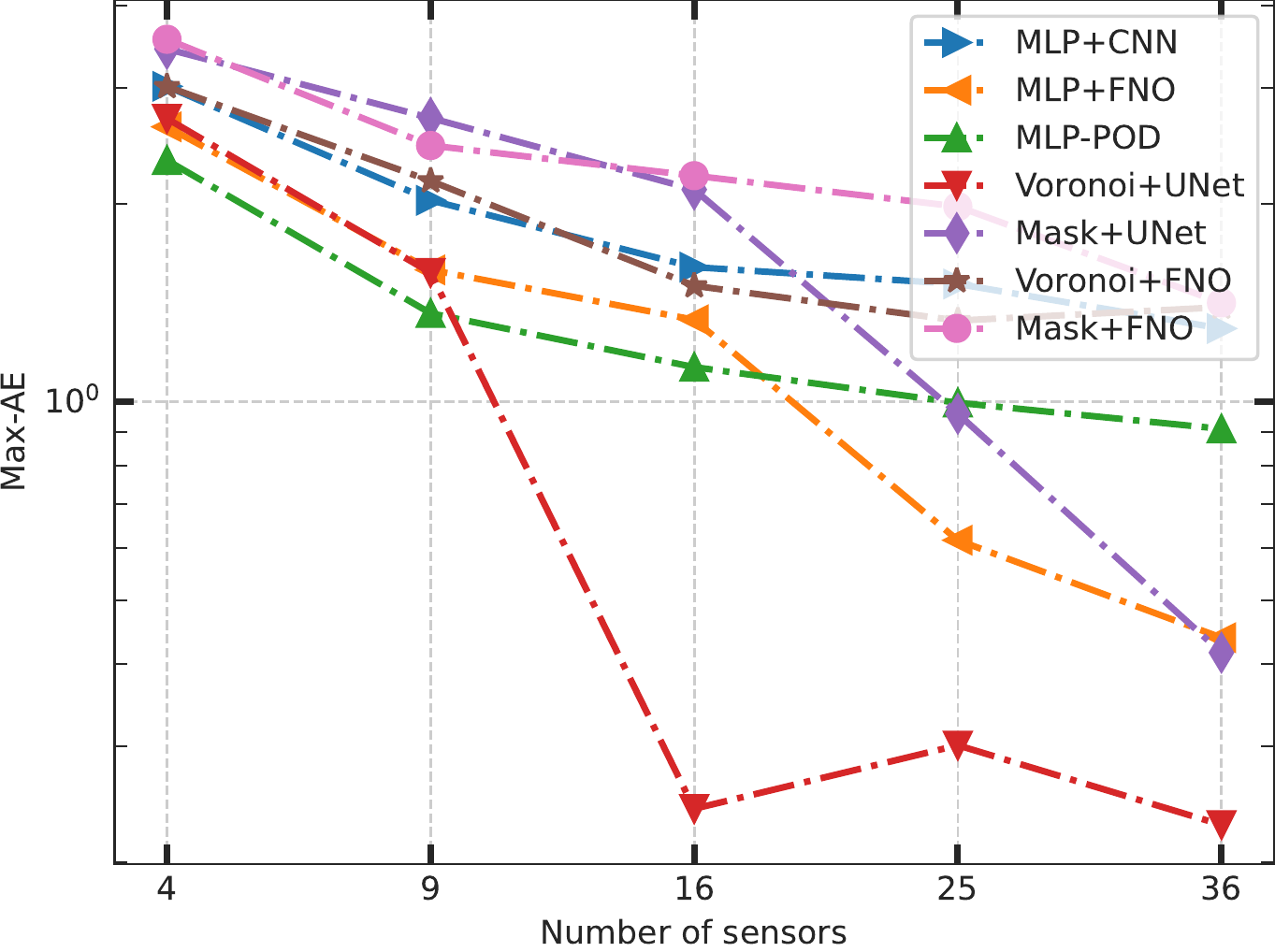}
	}
	\caption{Comparison of different approaches with an increasing number of sensors. All results are evaluated on Max-AE metric.}
	\label{fig:max-ae}
\end{figure*}

Max-AE metric evaluates the reconstruction error in the worst situation. As shown in Fig.~\ref{fig:max-ae}, the Max-AE metric follows a similar trend to MAE. However, it is different from MAE metric that MLP-POD performs well in sea surface temperature. 
As sea surface temperature is a realistic case where data is collected from real world, the temperature field is more complex and not noise-free. Leveraging the dictionary of POD modes to reconstruct physical field, MLP-POD avoids making predictions that are an outlier for historical data.   
In addition, the maximum error of FNO on heat conduct problem is larger than other methods. Since the heat is dissipated from the hole in the bottom, the temperature presents dramatic changes around the heat sink. The Fourier transform is hard to describe the drastic change in local, and the maximum reconstruction error appears near the hole, shown in Fig.~\ref{fig:heat_visual}. Apart from both sides of the holes, FNO is excellent for reconstructing the overall temperature distribution.

\subsubsection{Results visualization}
\begin{figure*}
	\centering
	\includegraphics[width=0.90\linewidth]{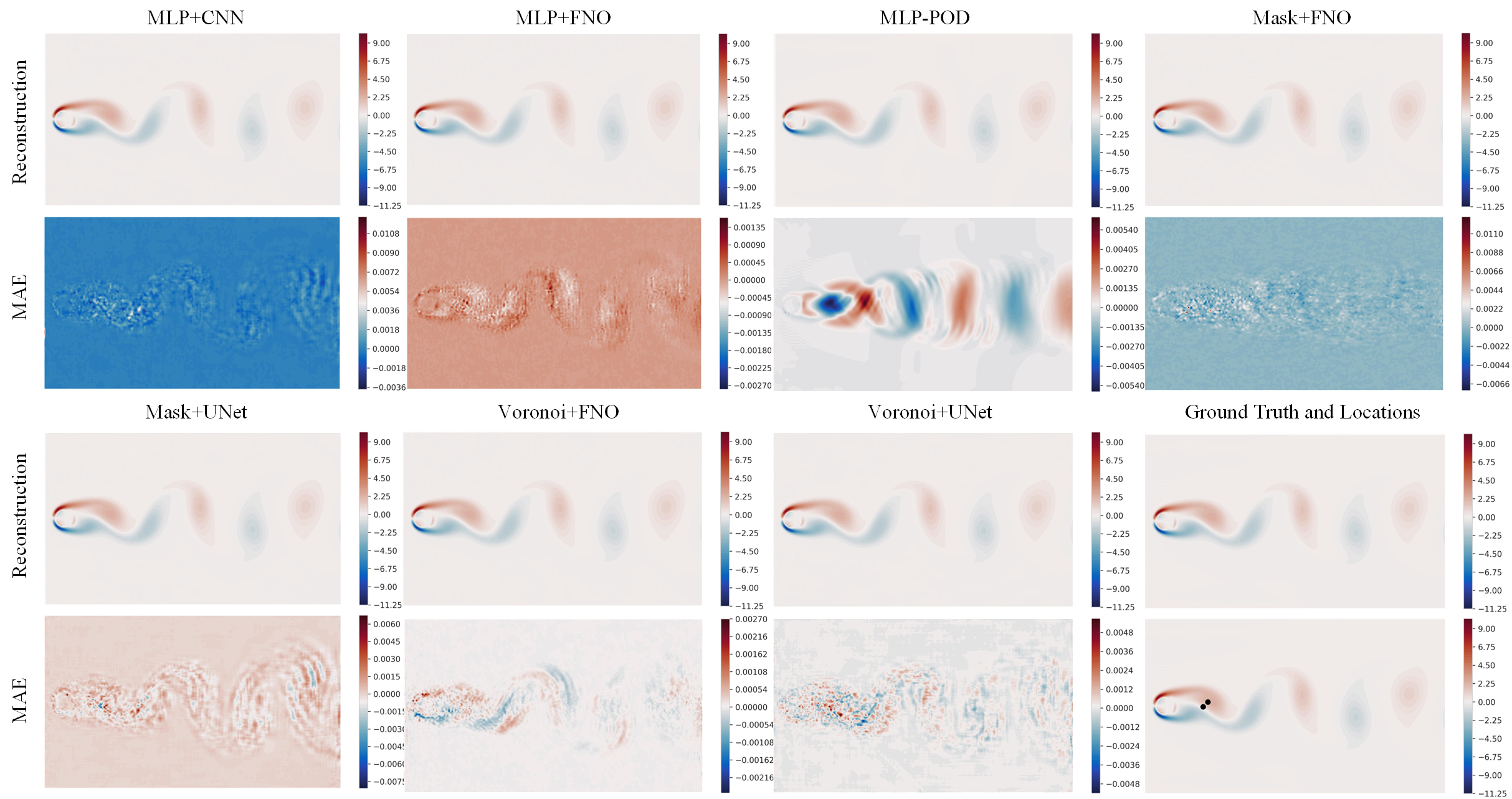}
	\caption{Visualization of reconstructed vorticity and absolute reconstruction error from 2 observations on cylinder wake dataset. The black dots on the last image represent locations of sensor.}
	\label{fig:cylinder_pre}
\end{figure*}

Fig.~\ref{fig:cylinder_pre} shows the reconstructed vorticity field and error maps on cylinder wake dataset. The proposed methods with different types of embedding are able to recover the vorticity accurately from 2 observations. In contrast, other methods present less accurate results. From the visualization of error maps, the reconstructed architecture based on FNO outperforms CNN and POD methods. The maximum reconstructed error of FNO architecture with MLP and Voronoi embedding is smaller than other methods. In addition, POD is a linear technology, and the error map of MLP-POD differs from FNO methods, directly learning the nonlinear mapping from observations to vorticity field. 

\begin{figure*}[t]
	\centering
	\includegraphics[width=0.80\linewidth]{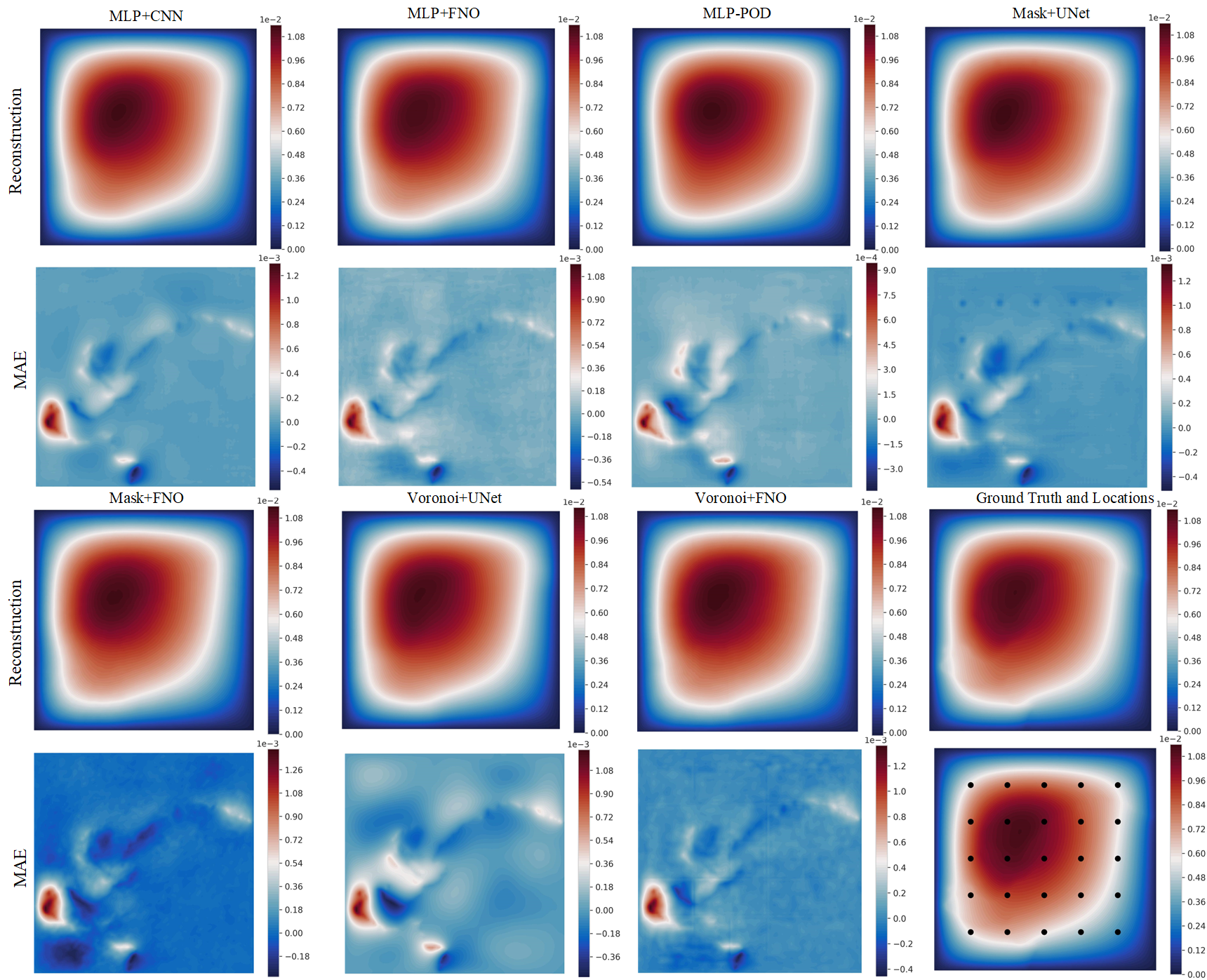}
	\caption{Visualization of reconstructed field and absolute reconstruction error from 25 observations on Darcy flow dataset. The black dots on the last image represent locations of sensor.}
	\label{fig:darcy_visual}
\end{figure*}

\begin{figure*}
	\centering
	\includegraphics[width=0.90\linewidth]{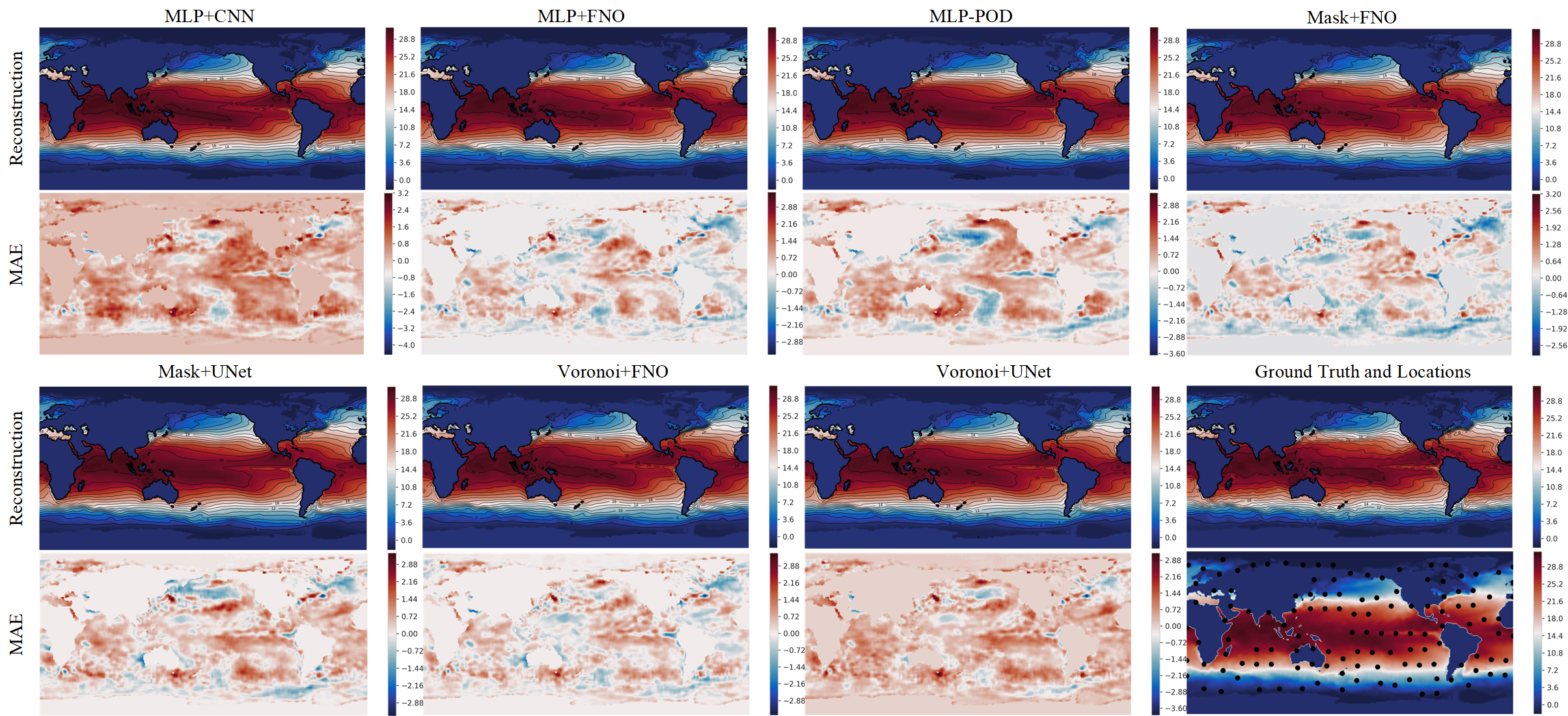}
	\caption{Visualization of reconstructed temperature field and absolute reconstruction error from 128 observations on SST dataset. The black dots on the last image represent locations of sensor.}
	\label{fig:sst_visual}
\end{figure*}

The visualization of results on Darcy flow dataset is shown in Fig.~\ref{fig:darcy_visual}. The physical field can be precisely reconstructed with the proposed method. Compared to cylinder wake dataset, the Darcy flow problem is a steady-state problem in which the variation among fields is not large. There is no significant difference between different methods, and the distribution of reconstructed error is similar. Fig.~\ref{fig:sst_visual} presents the visualization of results on SST dataset. It is a realistic and complex case that more observations are required to recover the temperature field. The visualization shows that the proposed method can obtain more competitive results than existing methods. It is noticeable that the proposed method is slightly better than POD and CNN-based methods from the isothermal line of reconstructed temperature field.  

\begin{figure*}[t]
	\centering
	\includegraphics[width=0.80\linewidth]{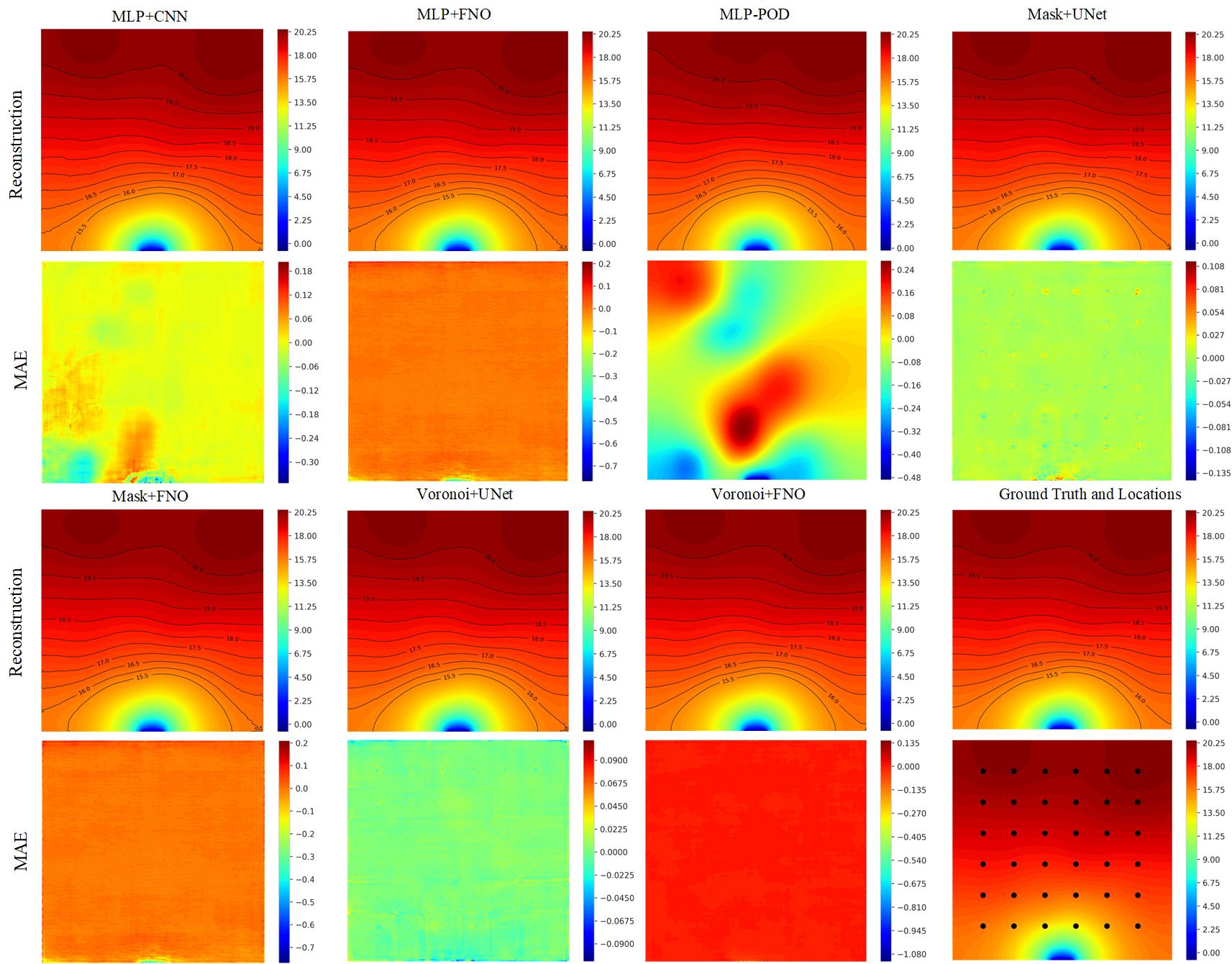}
	\caption{Visualization of reconstructed temperature field and absolute reconstruction error from 25 observations on heat conduction dataset. The black dots on the last image represent locations of sensor.}
	\label{fig:heat_visual}
\end{figure*}

Fig.~\ref{fig:heat_visual} shows the reconstructed steady-state temperature field on heat conduction dataset. The MLP-POD method is obviously worse than other methods from the isothermal line. In this case, the larger reconstructed error mainly appears around the heat sink on the bottom, and the proposed method performs relatively worse from the view of maximum error. However, the maximum error is only located on both sides of heat sink, and the average absolute error of FNO is better than other methods. The proposed method aims to recover the physical field in Fourier space. Since the Fourier transform is weak at processing abrupt signals, the FNO architecture is hard to precisely recover the local region with abrupt temperature changes around heat sink.

\subsection{Further Discussion}

\subsubsection{Zero-shot super-resolution reconstruction}

\cite{li2020fourier} has shown that Fourier neural operator is mesh-invariant to learn the mapping from functional parametric dependence to PDE solution. In our work, Fourier layers are leveraged to reconstruct the physical field in Fourier function space, so the proposed architecture is able to transfer to a higher resolution without retraining on higher resolution data. To demonstrate the capacity of zero-shot super-resolution reconstruction, we first train the model on a lower resolution and directly test the model on a higher resolution. Specifically, the trained model takes the higher-resolution input obtained by nearest neighbor interpolation, and produces the reconstructed physical field with the same resolution of inputs.

\begin{figure*}[tp]
	\centering
	\includegraphics[width=0.70\linewidth]{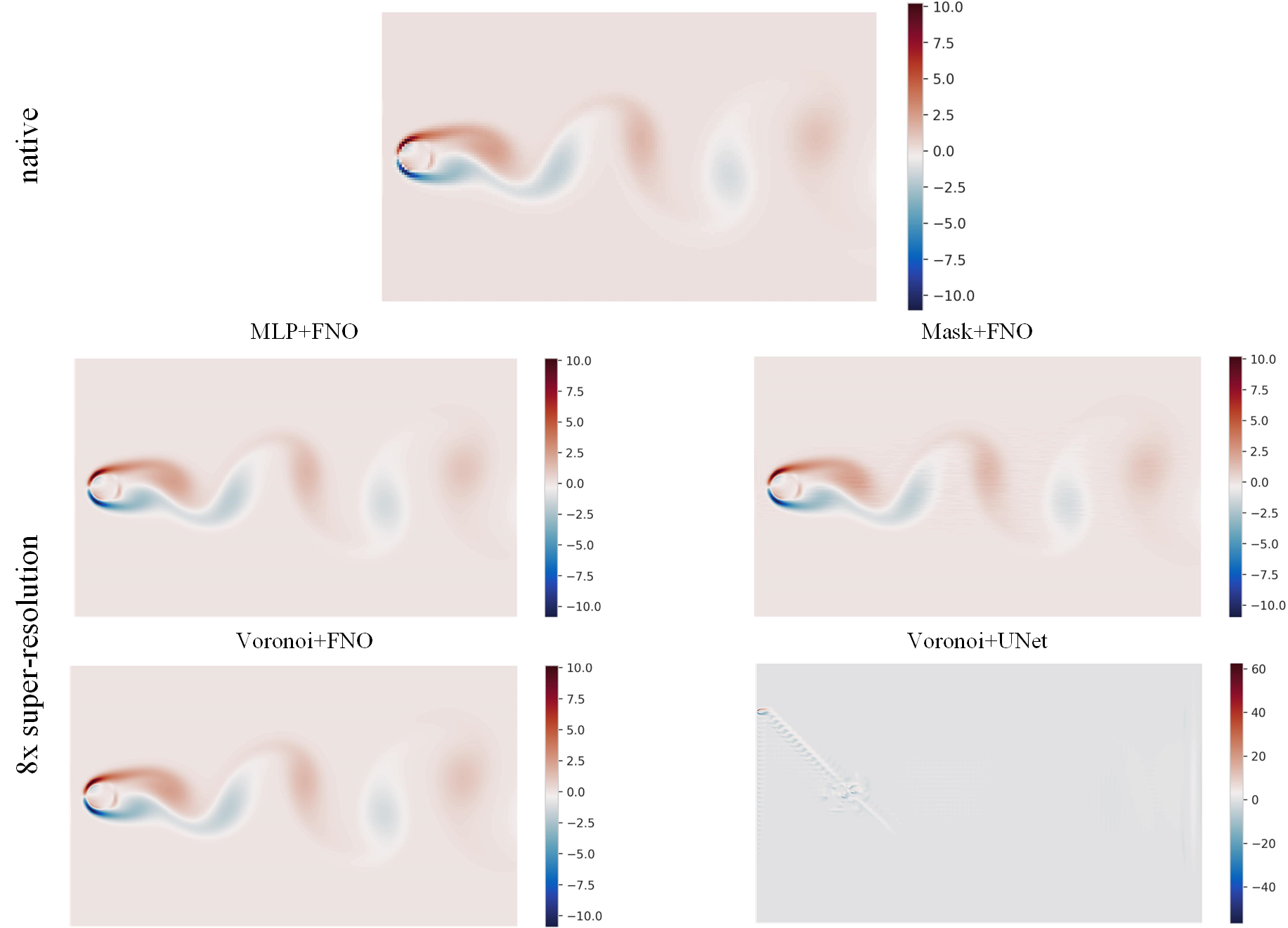}
	\caption{Illustrations of zero-shot super-resolution reconstruction on cylinder wake dataset.}
	\label{fig:cylinder_superresolution}
\end{figure*}

\begin{figure*}[tp]
	\centering
	\includegraphics[width=0.70\linewidth]{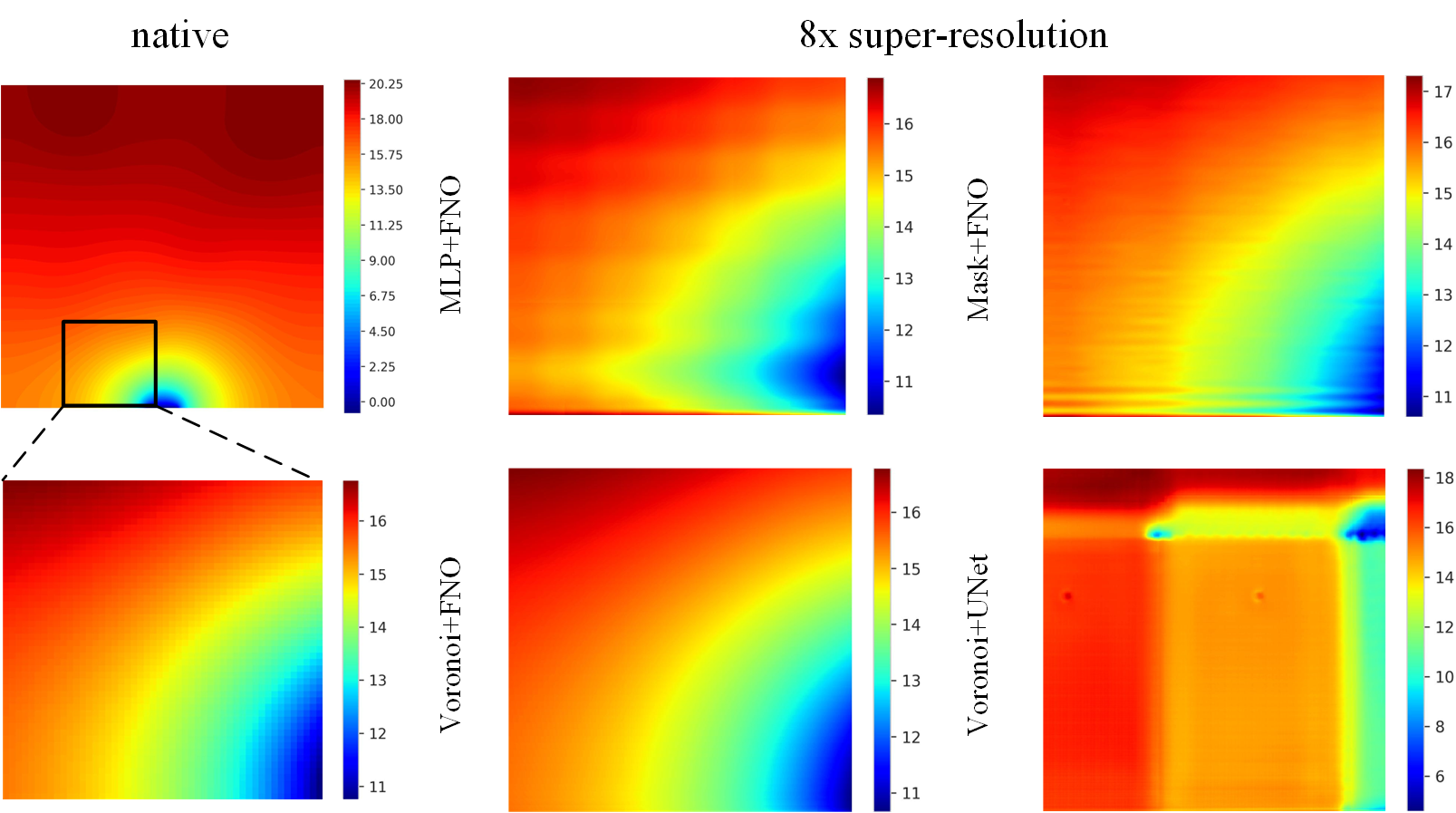}
	\caption{Illustrations of zero-shot super-resolution reconstruction on heat conduction dataset.}
	\label{fig:heat_superresolution}
\end{figure*}

Fig.~\ref{fig:cylinder_superresolution} shows the super-resolution results on cylinder wake dataset. The model is trained on 112$\times$192 resolution data and produces 896$\times$1536 resolution results, which is 8 times super-resolution on original data. This visualization demonstrates that the proposed method is able to achieve zero-shot super-resolution with various input representations. In contrast, the UNet has poor transferability among different meshes and is even unable to reconstruct coarse vorticity filed. As shown in Fig.~\ref{fig:heat_superresolution}, we present the super-resolution results of partial region on the left side of heat sink. For this example, the Voronoi embedding outperforms MLP and mask embedding on the super-resolution. After the Fourier transform, The inputs determine the initial point on the Fourier space. In this case, the Voronoi provide stable representations when interpolating the inputs to higher resolution, and mask and MLP cause inconsistent initial representations on Fourier space. Similar to cylinder wake case, the UNet cannot reconstruct temperature field on a higher resolution.          

The super-resolution results of MLP-POD are not compared because POD-based methods depend on the resolution of POD modes, and different resolution results with training data cannot be reconstructed.
Overall, benefiting from learning the reconstruction mapping in function space, the proposed method is mesh-invariant in contrast with existing reconstruction methods. The proposed method has the capacity to achieve zero-shot super-resolution. This characteristic is promising for leveraging multi-fidelity data with variable resolution.    

\subsubsection{Noise Addition}

\begin{figure*}[tp]
	\centering
	\subfigure[observations and physical field with noise]{
		\label{fig:noise_addition1}
		\includegraphics[width=0.36\linewidth]{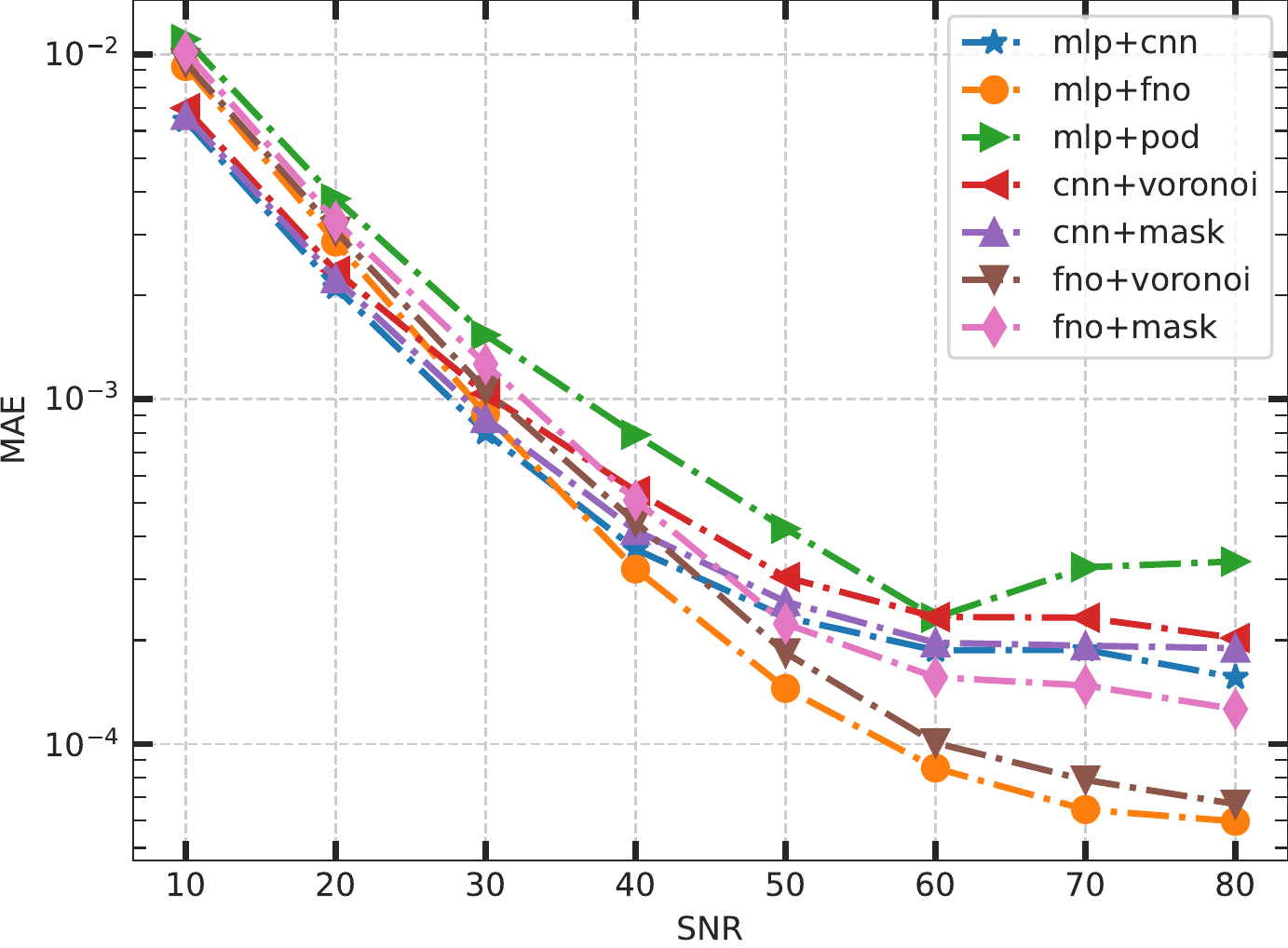}
	}
	\subfigure[observations with noise]{
		\label{fig:noise_addition2}
		\includegraphics[width=0.36\linewidth]{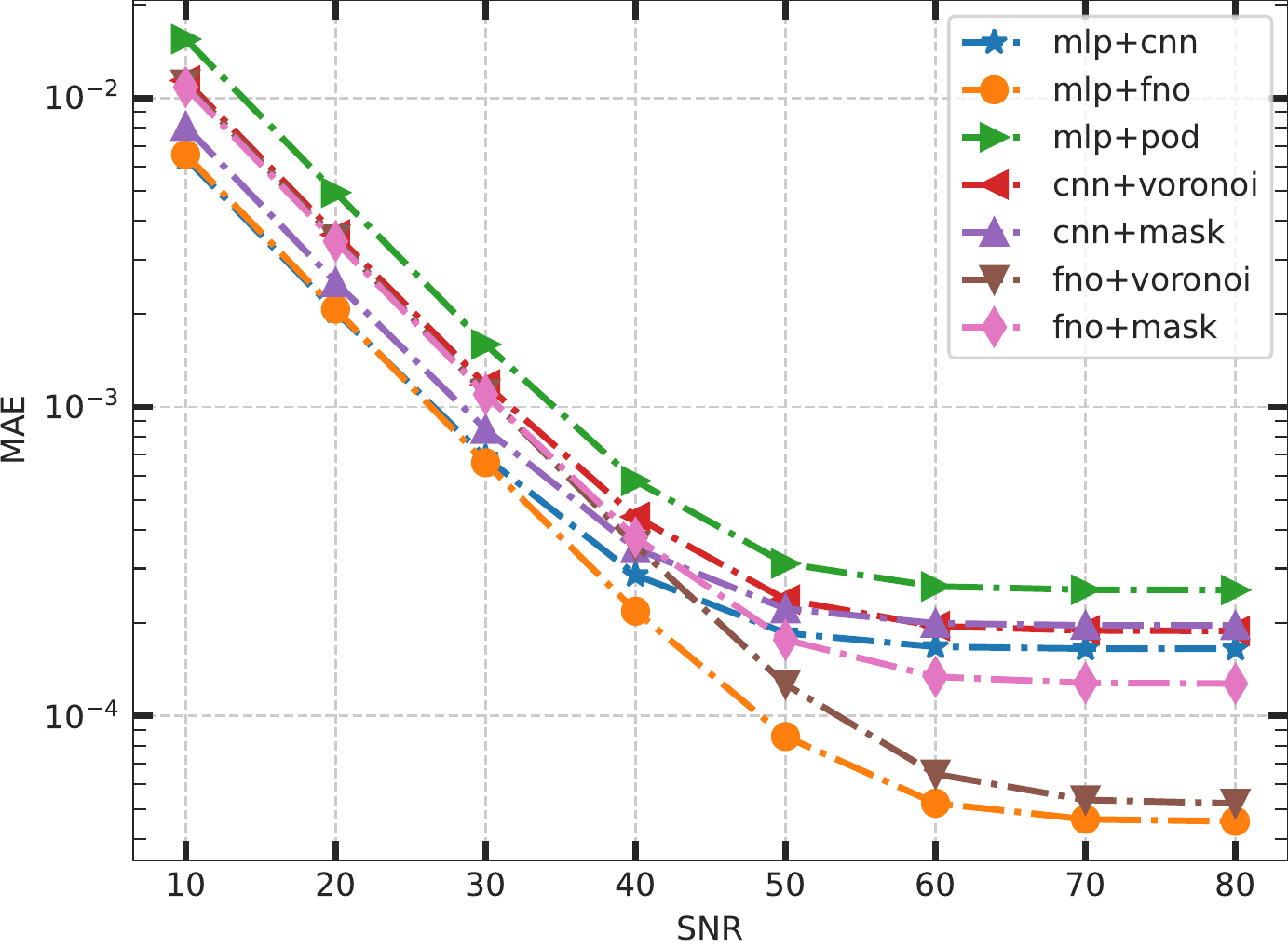}
	}
	\caption{Influence of different levels of noise applied on the obtained physical field.}
	\label{fig:noise_addition}
\end{figure*}

\begin{figure*}[tp]
	\centering
	\includegraphics[width=0.80\linewidth]{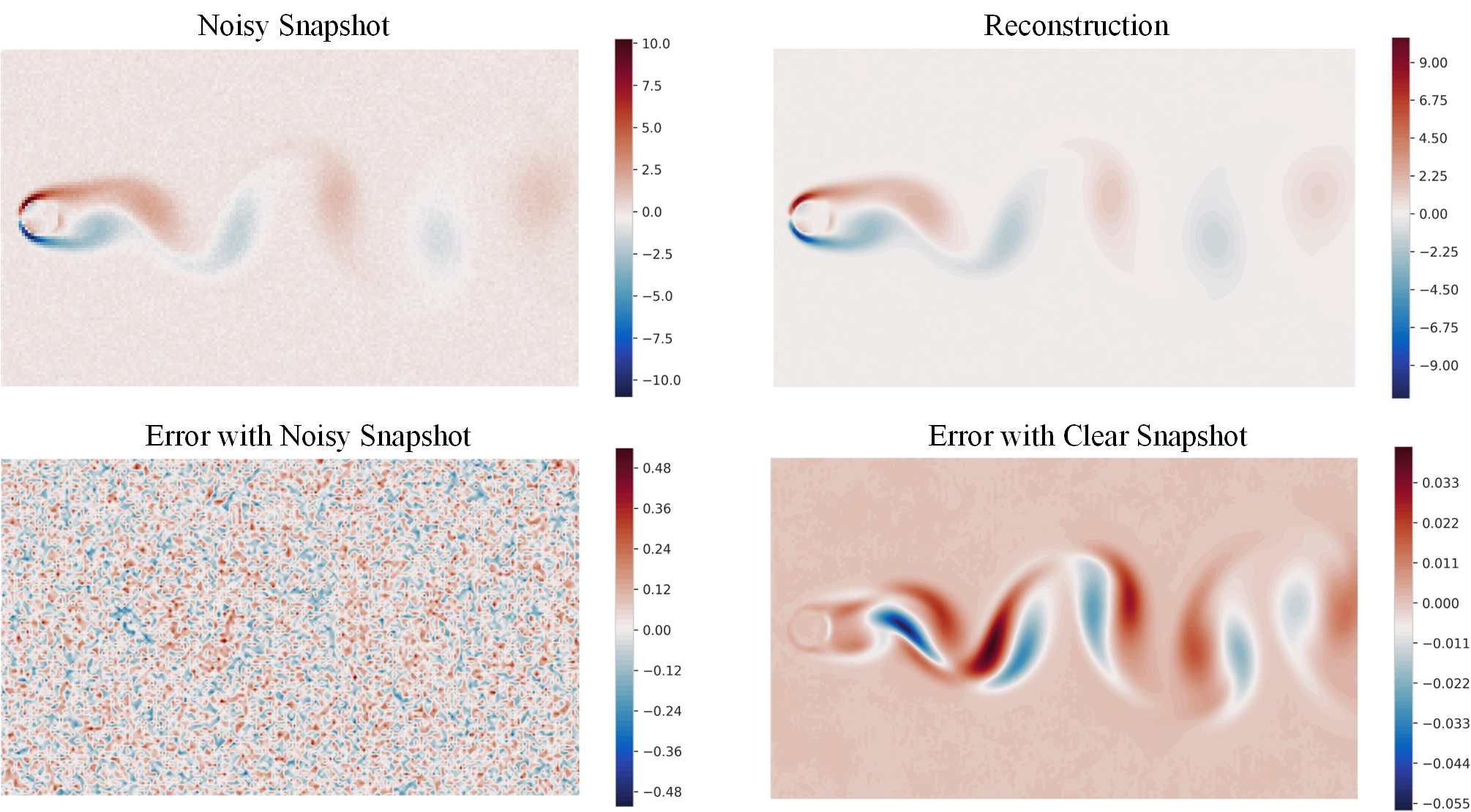}
	\caption{Visualization of the results using noisy snapshots. The noise level is 10 SNR.}
	\label{fig:noise}
\end{figure*}

Ideally, physical simulation is employed to generate training data with noise, and the observations from application environment are also noise-free. 
To investigate the robustness of proposed method in a more realistic setting, we consider two settings with the presence of noise.
In the first setting, the snapshots for training are supposed to be obtained in experimental studies with noise, so the noise exists together in the inputs and labels. 
Differently, the other setting expects that the reconstructed model is trained using simulated snapshots without noise. However, the model is evaluated by noisy inputs since there is noise in the observations measured in a realistic environment.     
In this section, we consider Gaussian white noise and use SNR to describe different levels of noise, which is formulated as:
\begin{equation}
	{\bf{u'}} = {\bf{u}} + {\left( {\frac{{\left\| {\bf{u}} \right\|_2^2}}{{{n_{\bf{u}}}}} \cdot \frac{1}{{2 \cdot {{10}^{\frac{{SNR}}{{10}}}}}}} \right)^{0.5}} \cdot \mathcal{N},
\end{equation}
where $\bf{u}$ and $\bf{u'}$ are respectively the noise-free and noisy label, and $\mathcal{N}$ is a random variable with standard normal distribution.

The effect of noise applied on inputs and outputs is shown in Fig.~\ref{fig:noise_addition1}. The performance of different methods is close in the low SNR range, and the CNN architecture yields slightly better results. In most SNR ranges, the proposed methods achieve better results. 
The results indicate that FNO is robust to noise, similar to CNN architecture, and can yield highly accurate results concurrently.
Fig.~\ref{fig:noise_addition2} presents the influence of setting where observations are corrupted by white Gaussian noise. Results obtained by the proposed method are better than other methods, especially for the FNO architecture with MLP embedding.
In addition, we observe that Voronoi is more sensitive to the presence of noise in observations than MLP embedding.
Fig.~\ref{fig:noise} shows the visualization of reconstructed vorticity field at low SNR. Markedly, the reconstruction is clearer than training snapshots corrupted by strong noise. It is supposed that the low-pass filtering in Fourier layer naturally suppresses noise. Additionally, reconstructing physical field in Fourier space can regularize the approximation of neural networks for more smoothness.

\subsubsection{Influence of Low-pass Filtering}

\begin{figure*}[tp]
	\centering
	\subfigure[MAE on cylinder wake]{
		\includegraphics[width=0.325\linewidth]{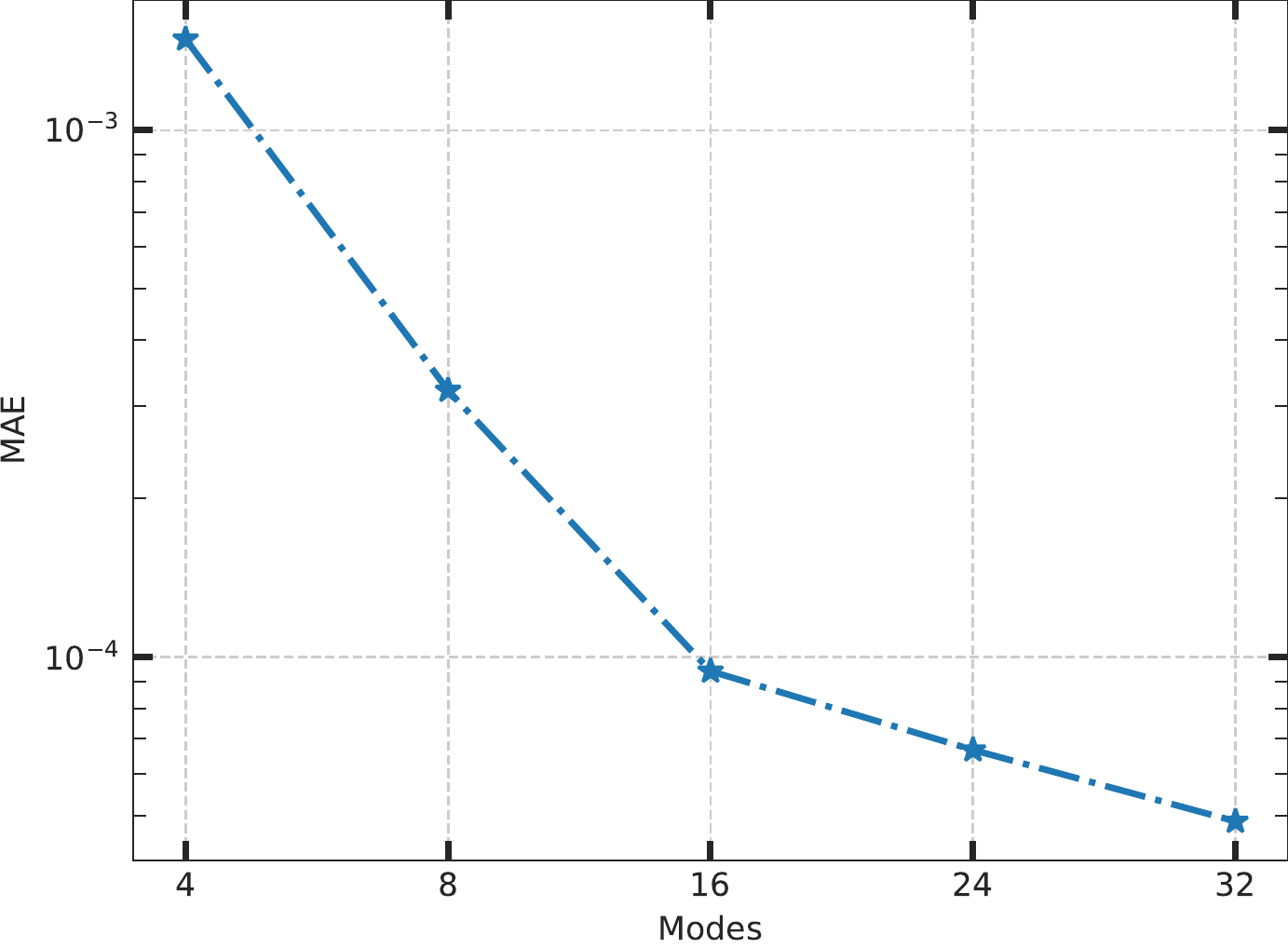}
		\label{fig:low_pass1}
	}
	\subfigure[Max-AE on cylinder wake]{
		\includegraphics[width=0.325\linewidth]{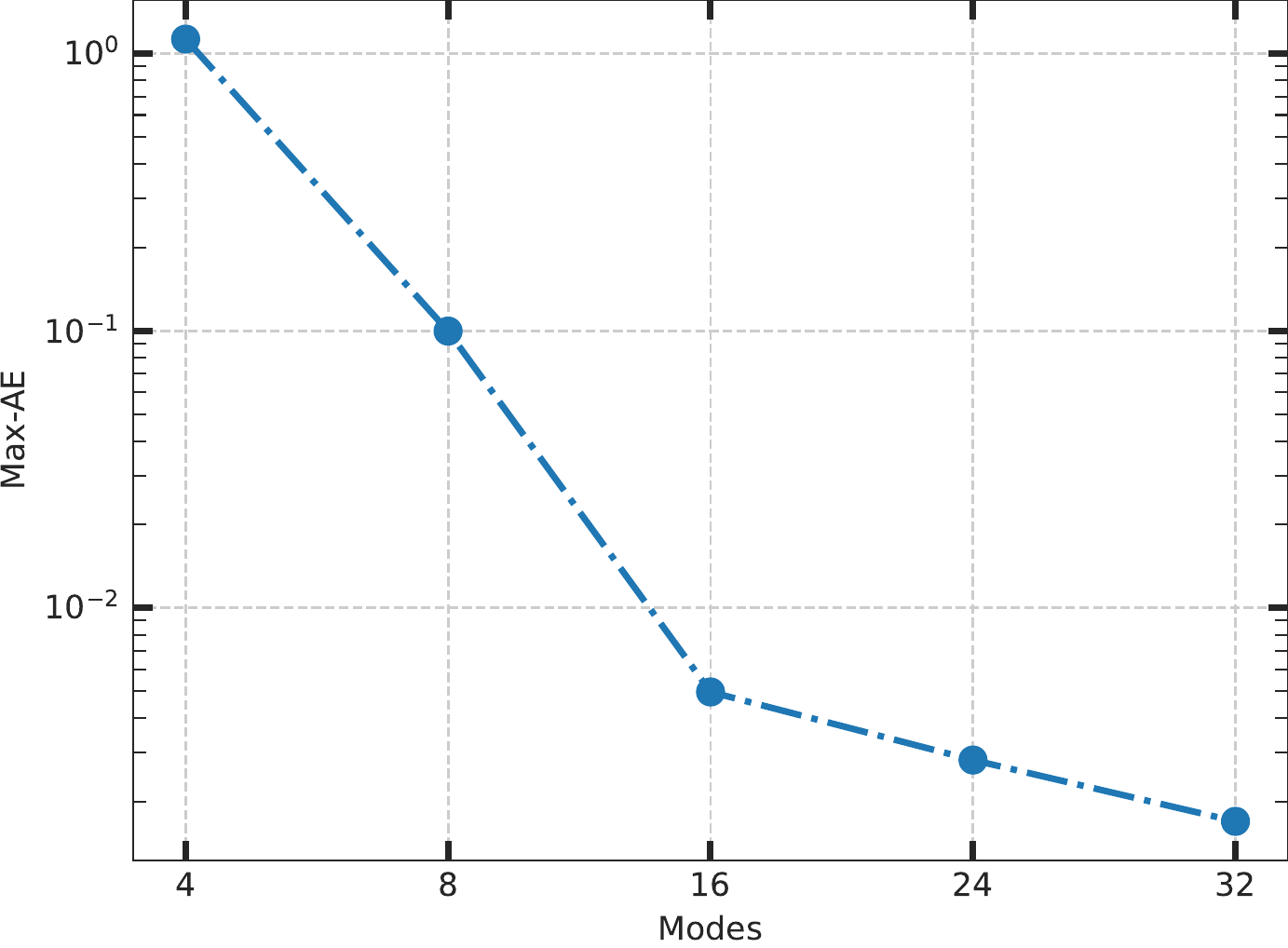}
		\label{fig:low_pass2}
	} \\
	\subfigure[MAE on heat conduction]{
		\includegraphics[width=0.325\linewidth]{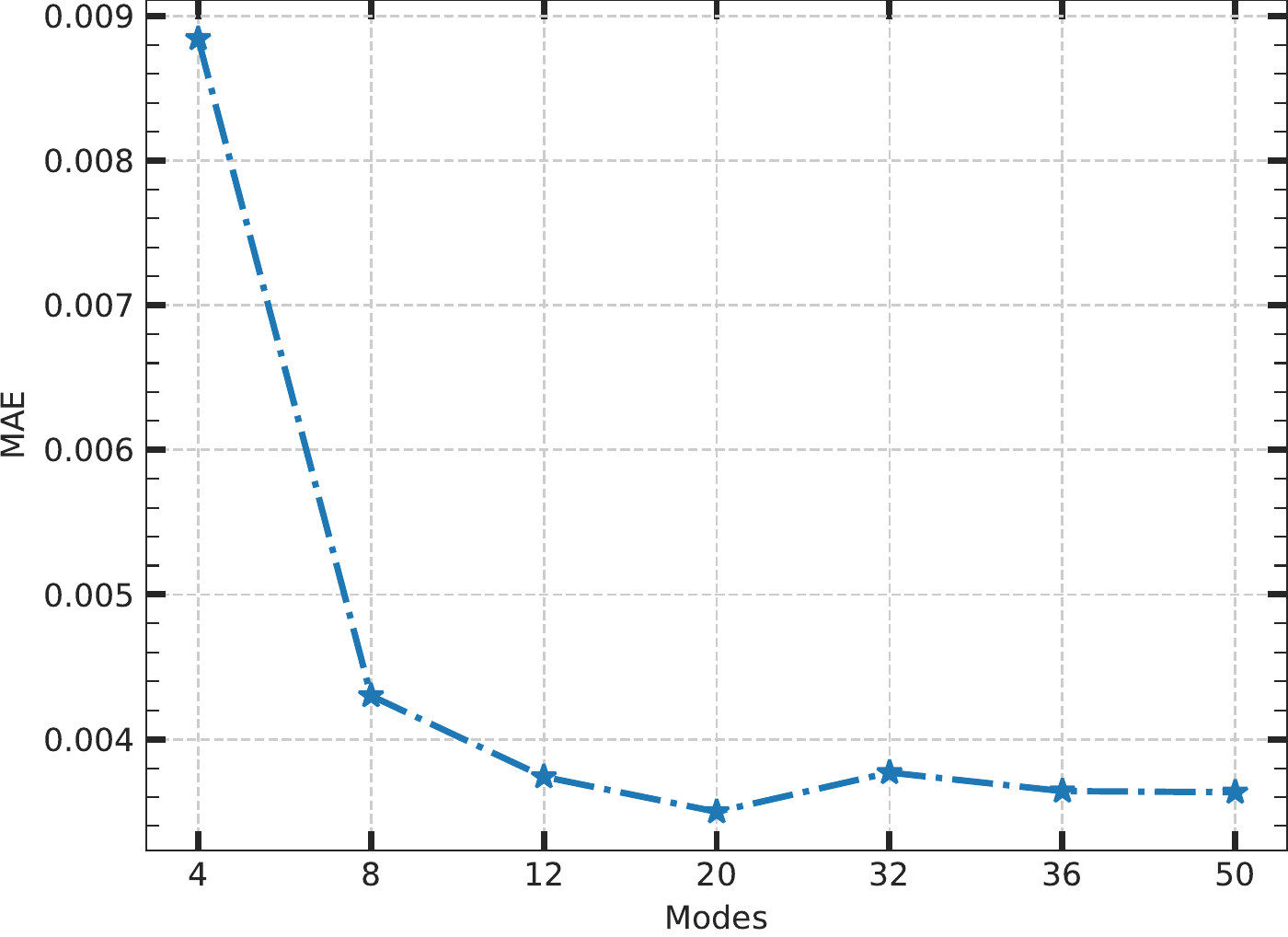}
		\label{fig:low_pass3}
	}
	\subfigure[Max-AE on heat conduction]{
		\includegraphics[width=0.325\linewidth]{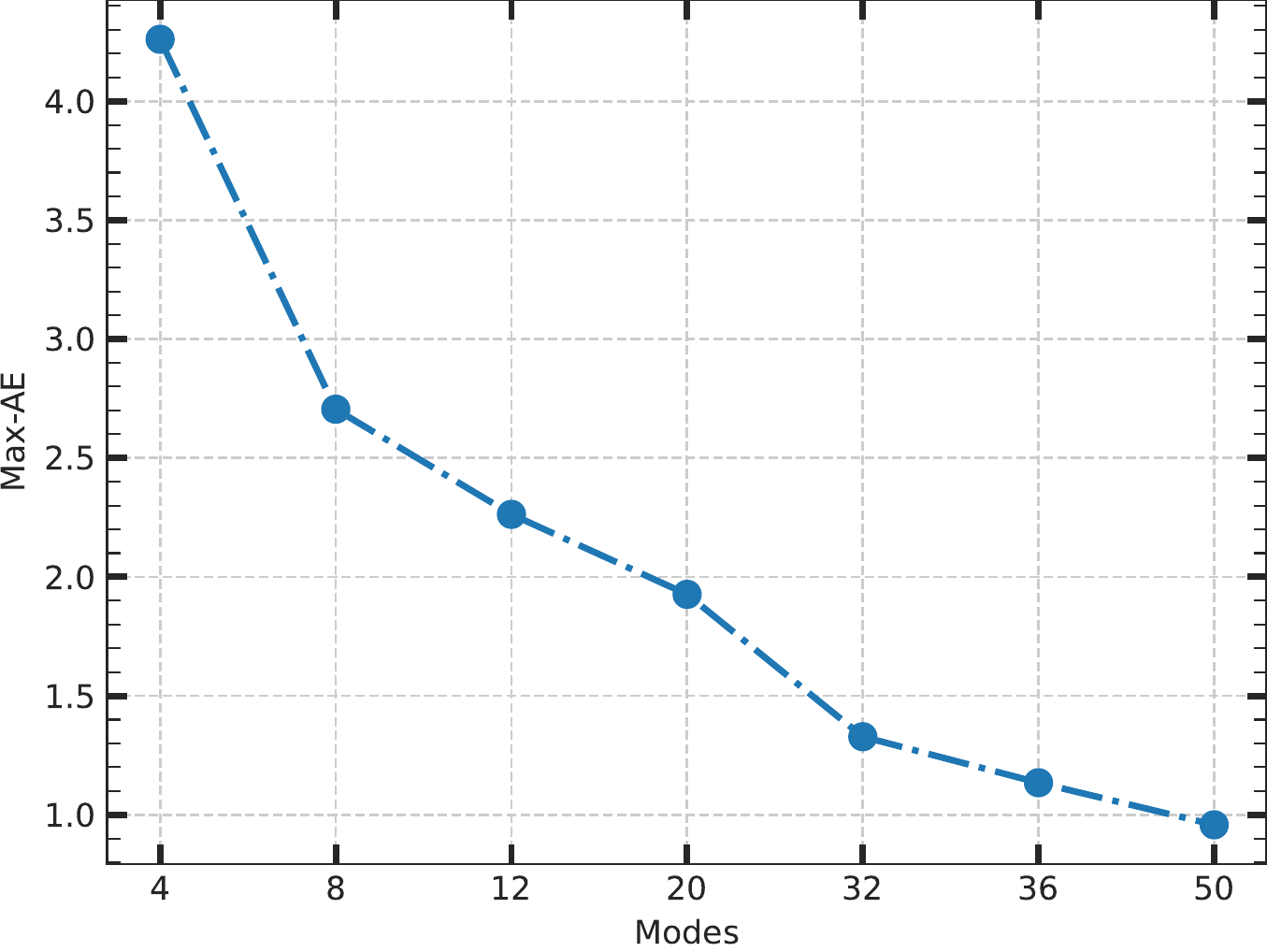}
		\label{fig:low_pass4}
	}
	\caption{Effect of the number of Fourier modes in low-pass filtering.}
	\label{fig:low_pass}
\end{figure*}

The $k_{max}$ Fourier modes are leveraged in each Fourier layer to characterize the underlying function for efficiency. In order to investigate the effect of low-pass filtering, ablation experiments are performed in this section.
Fig.~\ref{fig:low_pass} show the MAE and Max-AE metrics with various numbers of Fourier modes. 
When increasing the number of modes, the proposed method yields more accurate results on cylinder wake dataset.
In contrast, the improvement of MAE on heat conduction is insignificant when the number of Fourier modes is more than 10, while the Max-AE drops continuously because more modes are allowed to approximate the sharp change around heat sink.
It means the approximated capacity of FNO is improved at the expense of more parameters and computation, but the tradeoff should be made between accuracy and efficiency. 
 
\section{Conclusion}
\label{sec:5}
In this work, we propose a novel end-to-end method to reconstruct flow and heat field from sparse observations in infinite-dimensional space. It is a extension of neural operator rather than existing architecture that learns mappings between Euclidean spaces.
To appropriately represent the sparse observation data under various settings, we design three types of embedding for the proposed method, consisting of MLP, mask, and Voronoi embedding.
Secondly, Fourier neural operator is introduced to learn the mapping from sparse observations to global physical field in Fourier space. 
We expect to utilize function prior to regularize the reconstruction and obtain a more powerful nonlinear fitting ability. 
Additionally, benefiting from the resolution-invariant of Fourier layer, the proposed method has better transferability among different meshes.
Experiments on fluid mechanics and thermology problems demonstrate the feasibility and superiority of learning mapping in infinite-dimensional space. In most cases, the proposed method yields more accurate results than CNN and POD-based methods and achieves zero-shot super-resolution. It is a competitive architecture combining high precision reconstruction and excellent resolution transferability.

As a data-driven method, the performance of this work depends on the collected data. Future work pursues to develop the physics-informed neural network to solve reconstruction problems with less training data and high generalization. In general, the domain of physical field is irregular. It is promising to extend existing deep learning-based methods for physical field reconstruction to the more complex computation domain.  

\section*{Acknowledgment}
This work was supported by the National Natural Science Foundation of China (No.11725211 and 52005505).

\section*{CRediT authorship contribution statement}
\textbf{Xiaoyu Zhao:} Writing - original draft, Methodology, Implementation, Investigation, Data generation.
\textbf{Xiaoqian Chen:} Writing - review \& editing, Methodology, Project administration, Supervision.
\textbf{Zhiqiang Gong:} Writing - review \& editing, Methodology.
\textbf{Weien Zhou:} Writing - review \& editing, Methodology, Supervision.
\textbf{Wen Yao:} Writing - review \& editing, Methodology, Project administration, Supervision.
\textbf{Yunyang Zhang:} Writing - review \& editing, Methodology, Visualization.

\section*{Declaration of competing interest}
The authors declare that they have no known competing financial interests or personal relationships that could have appeared to
influence the work reported in this paper.

\bibliography{mybibfile}

\end{document}